\documentclass[10pt,twocolumn,letterpaper]{article}

\usepackage[pagenumbers]{cvpr} 

\usepackage[dvipsnames]{xcolor}

\definecolor{cvprblue}{rgb}{0.21,0.49,0.74}
\usepackage[pagebackref,breaklinks,colorlinks,citecolor=cvprblue]{hyperref}

\usepackage{algorithm,algorithmic}
\usepackage{amsmath}

\renewcommand{\algorithmiccomment}[2][.4\linewidth]{%
  \leavevmode\hfill\makebox[#1][l]{//~#2}}

\def\paperID{3992} 
\def\confName{CVPR}
\def\confYear{2024}

\title{Visual Hindsight Self-Imitation Learning for Interactive Navigation}

\author{Kibeom Kim, Kisung Shin, Min Whoo Lee, Moonhoen Lee\\
Seoul National University\\
{\tt\small \{kbkim, ksshin, mwlee, mhlee\}@bi.snu.ac.kr}
\and
Minsu Lee\thanks{Corresponding authors.}, Byoung-Tak Zhang$^\ast$\\
Seoul National University, AIIS\\
{\tt\small \{mslee, btzhang\}@bi.snu.ac.kr}
}

\begin{document}
\maketitle
\begin{abstract}
Interactive visual navigation tasks, which involve following instructions to reach and interact with specific targets, are challenging not only because successful experiences are very rare but also because the complex visual inputs require a substantial number of samples.
Previous methods for these tasks often rely on intricately designed dense rewards or the use of expensive expert data for imitation learning. 
To tackle these challenges, we propose a novel approach, Visual Hindsight Self-Imitation Learning (VHS) for enhancing sample efficiency through hindsight goal re-labeling and self-imitation.
We also introduce a prototypical goal embedding method derived from experienced goal observations, that is particularly effective in vision-based and partially observable environments. 
This embedding technique allows the agent to visually reinterpret its unsuccessful attempts, enabling vision-based goal re-labeling and self-imitation from enhanced successful experiences. 
Experimental results show that VHS outperforms existing techniques in interactive visual navigation tasks, confirming its superior performance and sample efficiency. 
\end{abstract}

\section{Introduction}
\label{sec:intro}

Embodied AI \cite{duan2022survey,savva2019habitat,franklin1997autonomous,xia2018gibson,das2018embodied}, which focuses on enabling artificial intelligence to sense, understand, and act in a human-like manner, is rapidly evolving and gaining significance.
It can assist people in a variety of ways. 
For instance, it is increasingly being applied in service robots for tasks like indoor errands, as demonstrated in several studies \cite{yi2019mobile,ramalingam2020human,lee2020visual,zeng2021pushing,RoomR}.
In the context of interactive visual navigation tasks \cite{wortsman2019learning,zeng2021pushing,al2022zero,kim2021goal,wu2018building,RoomR}, an agent must be able to perceive and understand its surroundings from a first-person perspective and follow given instructions by word embedding.
However, these tasks demand a vast amount of samples for learning, considering the high-dimensional visual inputs.
In addition, useful feedback is only sparsely provided in such environments, making sophisticated behaviors difficult to attain through random exploration.



\begin{figure*}[t]
  \centering
   \includegraphics[width=0.9\linewidth]{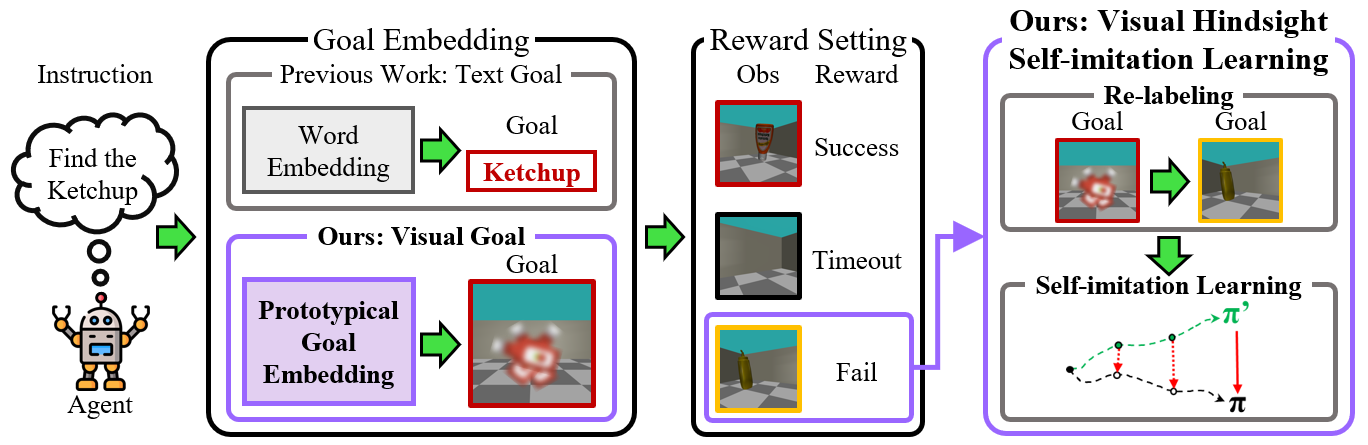}

   \caption{Illustration of the learning process for proposed method. This approach employs prototypical goal embeddings to pursue goals, diverging from traditional word-embedding instructions. It introduces a strategy for re-labeling goals in failed episodes. 
   The method facilitates learning through self-imitation and benefits from sparse reward settings in environments with visual inputs.}
   \label{fig:concept}
\end{figure*}


The visual navigation tasks require ten of millions to billions of samples \cite{wijmans2019dd,maksymets2021thda,mezghan2022memory} to reliably reach a specified goal.
For these tasks, various solutions have been proposed, such as designing reward functions or imitating the behavior of experts.
Among them, studies that utilize dense reward functions \cite{zeng2021pushing,wu2018building,wortsman2019learning,al2022zero,savva2019habitat,maksymets2021thda} require task-specific experts to design such complex rewards.
An alternative is imitation learning \cite{sun2018truncated,zhu2020off,hussein2017imitation,ho2016generative,schaal1999imitation,fang2022target,thakur2023imitation,karnan2022voila} that utilizes expert demonstrations, but the need to collect expensive datasets from experts remains a widely recognized drawback.
As a way to enhance exploration without expert data or complex reward shaping, methods based on Hindsight Experience Replay (HER) have been proposed \cite{andrychowicz2017hindsight,dai2021diversity,fang2019curriculum,manela2021bias,dai2020episodic,fang2018dher,tang2021hindsight}.
These methods tackle sparse-feedback, hard-exploration problems \cite{paine2019making,burda2018exploration,ecoffet2019go,andrychowicz2017hindsight,fang2020adaptive} such as joint control tasks by re-labeling the goals of failed trajectories during training, turning them into learnable experiences.
However, they assume fully observable environments and use numerical coordinates and robot poses as goals, which is unsuitable for partially observable environments with visual observations.
While there has been an effort to combine HER with vision input by introducing a generative model \cite{sahni2019addressing}, this requires collecting goal-related data and training the model in advance.
Thus, challenge remains in applying the idea of re-labeling failed episodes in vision-based, partially observable domains.


We propose the \textbf{V}isual \textbf{H}indsight \textbf{S}elf-Imitation Learning (\textbf{VHS}) to address the problems of sample efficiency and hard exploration in instruction-based visual interactive tasks.
The pivotal idea behind VHS is 1) to enhance sample efficiency by utilizing failed episodes by re-labeling desired goals or actions to achieved goals or actions and 2) to exploit past good experiences and promote exploration by self-imitation with enriched successful episodes.
To enable re-labeling, we devise Prototypical Goal (PG) embedding: instead of word embedding from instruction \cite{kim2021goal,kim2023sa,wu2018building}, we replace the given instruction with prototypical feature, which is the average of the collected features of the instructed goal as shown in Figure~\ref{fig:concept}.
While allowing the goal to be vision observation, this method also enables re-labeling in hindsight the failed trajectories with the goal and the interaction type the agent actually interacted with.
Our method allows the agent to learn from negative rewards in failed episodes and from self-imitation in successfully re-labeled episodes, facilitating efficient learning that does not rely on expert demonstrations. We achieve leading success rates and demonstrate high sample efficiency in challenging interactive visual navigation tasks.
We demonstrate state-of-the-art success rates and sample-efficiency on challenging interactive visual navigation tasks.
The visualization of our proposed prototypical goal embedding illustrates its effectiveness in encapsulating the visual characteristics of goals. 
Additionally, ablation studies reveal that it outperforms the standard word embedding in interactive visual navigation environments.

Our contribution points in this paper are as follows.\\
1.	We propose novel Visual Hindsight Self-imitation (VHS) learning method for interactive visual navigations. VHS enhances sample efficiency by utilizing hindsight to re-label failed episodes and promotes in-depth exploration through self-imitation learning with the enriched successful episodes.\\
2.	We introduce Prototypical Goal (PG) embeddings to enable re-labeling in vision-based, partially observable environments. It enables efficient vision-based re-labeling and performance improvement compared to word embeddings.\\
3.	Experimental results show state-of-the-art success rates and sample efficiency of our method demonstrating its effectiveness in challenging interactive navigation tasks.



\section{Related Work}
\label{sec:related_work}

\subsection{Visual Navigation}

Visual navigation tasks, aimed at finding a goal in indoor environments from a first-person perspective, have diversified into specialized categories such as scene-driven navigation, instruction-based visual navigation, and interactive tasks. 
Scene-driven navigation tasks \cite{zhu2017target,wu2020towards,fang2022target,lyu2022improving,devo2020towards,mezghan2022memory} involve guiding an agent to reach a specific or similar goal in the environment based on a given picture of the goal. 
Our method is inspired by these studies, particularly the concept of pursuing features from an image or a multi-modal \cite{al2022zero} goal, leading us to adopt prototypical goal embedding in the instructions. This enables the extraction and pursuit of desired goals. However, our approach uniquely leverages the agent’s own experiences of goal stages during learning, as opposed to using a given target image.

Instruction-based visual navigation is receiving increasing attention \cite{wu2018building,kim2021goal,kim2023sa,chaplot2020object,savva2019habitat,maksymets2021thda}.
In these studies, the instruction specifies the desired goal among various targets in the environment, with a focus on exploring and recognizing targets \cite{wu2018building,kim2021goal,kim2023sa}.
Additionally, there are tasks \cite{zeng2021pushing,al2022zero,RoomR,xiang2020sapien,ai2thor} that perform interaction with various goals based on interactive indoor environments.
However, due to the difficulty of the task, which requires reaching the goals in visual navigation and performing the appropriate interaction in interactive navigation, they rely on pre-trained models for identifying targets and sophisticatedly designed dense rewards for efficient exploration. 
On the other hand, our approach uses straightforward and intuitively defined rewards of \{Success, Timeout, Timestep penalty, and Fail\} and can interact with a wide variety of target objects without the need for pre-trained models.

\subsection{Self-Imitation Learning with HER}


Self-imitation learning (SIL) \cite{oh2018self,li2023self,luo2021self,tang2020self,pshikhachev2022self,ghosh2020learning,dai2020episodic} is used for exploiting past under-estimated experiences to derive deep exploration indirectly. 
SIL involves storing experiences in a replay buffer and focusing on imitating state-action pairs in the replay buffer only when the return in the past episode is greater than the agent’s estimated value. 
SIL can apply dense reward settings for estimating value and return.
SIL offers advantages in enhancing exploration effectively based on previous experiences, eliminating the need for expensive demonstration data. 
However, SIL might not perform well in environments with sparse rewards. 
Since it relies on past rewards to guide learning, the lack of frequent rewards can hinder its effectiveness.

In goal-conditioned Reinforcement Learning (RL), studies have combined SIL and HER as a way to increase the frequency of positive reward and improve sample-efficiency, even in sparse reward settings.
Episodic self-imitation learning with hindsight (ESIL) \cite{dai2020episodic} is a self-imitation algorithm that leverages entire episodes with hindsight whereas original self-imitation learning samples state-action pairs with underestimated value from the experience replay buffer. 
ESIL includes a trajectory selection module to filter uninformative samples from each episode of the update and an adaptive loss function. 
GRSIL \cite{li2023self} studies learn actor policies and self-imitated policies with goal re-labeling separately, and then combine them to infer the most rewarding actions. 
These studies are similar to ours in that they use the agent's past experience and re-labeling, but they have the limitation of learning in an environment where the goal coordinates and robot pose are fully observable.
Unlike these works, we propose to re-label and self-imitate experiences in vision-based, partially observable environments. 


\begin{figure*}[t]
  \centering
   \includegraphics[width=0.9\linewidth]{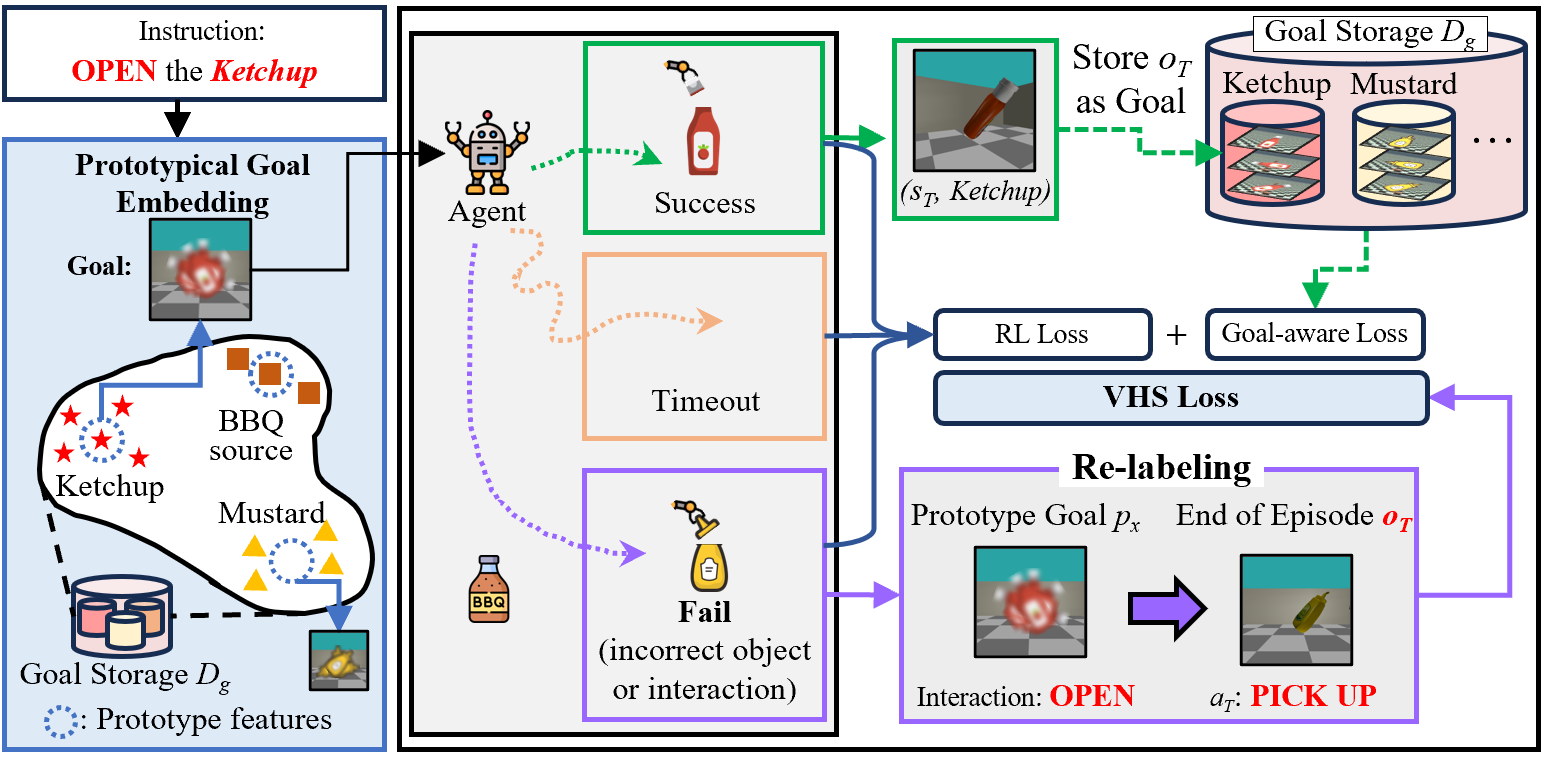}
   \caption{The overall architecture. 
   Prototypical Goal (PG) embedding samples goal observations collected from goal storage, extracts features using the feature extractor from agent, and computes prototypical features. 
   If the episode ends successfully, the goal observation is collected by pairing the end-of-episode observation with the desired goal. For each episode that fails, the re-labeling process is run to replace the goal with the last observation of the episode, and then perform the Visual Hindsight Self-Imitation Learning.}
   \label{fig:overall}
\end{figure*}

\section{Preliminaries}

\subsection{Reinforcement Learning}
The instruction-based interactive visual navigation tasks we consider, where the agent is given an RGB image from an egocentric camera view and an instruction from environment. For the tasks, we use the Partially Observable Markov Decision Process (POMDP) as a tuple  $\left(\mathcal{S}, \mathcal{O}, \mathcal{A}, \mathcal{R}, \mathcal{P}, \gamma, \mathcal{I}\right)$, where $\mathcal{S}$ denotes the state space, $\mathcal{O}$ the observation space,  $\mathcal{A}$ the action space, $\mathcal{R}$ the reward function, $\mathcal{P}$ the transition probability function, and $\gamma \in [0,1)$ the discount factor.
We note that in the interactive visual navigation tasks that we tackle, the action space is $\mathcal{A} = \mathcal{M} \cup \mathcal{K}$, where $\mathcal{M}$ is the set of actions for agent movements and $\mathcal{K}$ is the set of actions for agent's interaction with goal objects.
$\mathcal{I}$ denotes the set of \textit{instructions}, where each instruction $I^{k, x} \in \mathcal{I}$ specifies an interaction $k \in \mathcal{K}$ and a goal $x \in \mathcal{X}$ among possible interactions $\mathcal{K}$ and goal objects $\mathcal{X}$.
In each episode, an instruction $I^{k,x}$ specifies with which goal ($x$) and via which interaction type ($k$) the agent must interact.
As outlined later in Sec.~\ref{sec:rwd_settings}, the reward function is determined based on the given instruction.
State-value function in this instruction-based POMDP is $V(o, I^{k,x}) = V(o| k,x) = \mathbb{E}[R_{t}| O_t=o, X=x, K=k]$, where $O_t$ is observation at time $t$ and $X$, $K$ are goal and interaction type given by the instruction respectively.
$R_t = \sum_{t'=t}^{T} \gamma^{(t'-t)}r_{t'}$ denotes the sum of decayed rewards $r_{t'}$ from time step $t$ to terminal step $T$.
The aim of an agent is to find a policy $\pi : \mathcal{O} \rightarrow \Delta(\mathcal{A})$ that maximizes $R_t$.

As the base reinforcement learning algorithm, we use Asynchronous Advantage Actor-Critic (A3C) \cite{mnih2016asynchronous}.
In this method, the policy gradient for the actor function $\pi_\theta$ and the loss gradient for the critic function $V_\phi$ are defined as Eq. \ref{eq:actor} and \ref{eq:critic} respectively:
\begin{align}
    \nabla_\theta \mathcal{L}_{RL} =& -\nabla_\theta \log \pi_\theta (a_t|o_t, x, k)(R_t-V_\phi(o_t| x, k)) \nonumber\\
    &-\beta \nabla H(\pi_\theta(\,\cdot\,|o_t, x, k)) \label{eq:actor}\\
    \nabla_\phi \mathcal{L}_{RL} =& \, \nabla_\phi (R_t-V_\phi(o_t| x, k))^2 \label{eq:critic} 
\end{align}

\subsection{Goal-Aware Learning}

In visual navigation tasks, it is crucial for an agent to recognize goals based on its experiences. We use goal-aware representation learning \cite{kim2023sa,kim2021goal} to train the agent in distinguishing between different targets during policy learning. This involves using the SupCon \cite{khosla2020supervised} loss, a contrastive learning method that pairs same-labeled data as positives. Additionally, this approach requires datasets with labeled goal observations.
When an instruction is successfully executed, the observation at the terminal step is considered a \text{goal observation}, and a \textit{(goal observation, desired goal $x$)} pair is stored in the \textit{Goal Storage} $D_g$, which serves as the dataset for representation learning.
In this setup, the desired goals in the instructions act as labels for the goal observations. Data corresponding to the same target are treated as positive pairs, while data for different targets are negative pairs. This framework helps train the feature extractor using the Goal-aware SupCon loss function $\mathcal{L}_s$, which is defined as follows:
\begin{equation}
\label{eq:supcon}
\mathcal{L}_{s} = \sum_{j \in J} {-1 \over |P(j)|}\sum_{p \in P(j)} \log {\exp(g_j \cdot g_p / \tau_s)\over \sum\limits_{h \in J\setminus\{j\}} \exp(g_j \cdot g_h/\tau_s)}
\end{equation}
where $J$ is the set of indices of goal states in the batch, and $P(j)$ is the set of all positive pair indices corresponding to the $j$-th goal state (where $j$ itself is not in $P(j)$).
$|P(j)|$ is the cardinality of $P(j)$, $\cdot$ is the dot product operator, $g_j$ is the output of the feature extractor of $j$-th goal state, and $\tau_s$ is temperature as a hyperparameter.

\subsection{Reward Settings}
\label{sec:rwd_settings}
In order to perform interactive visual navigation tasks, it is common to design complex rewards based on relative distance \cite{wu2018building,zeng2021pushing,wortsman2019learning,savva2019habitat,maksymets2021thda} to the target or observation of agent \cite{al2022zero,wu2018building}.
To ensure that even simple and intuitive reward setting can perform the task correctly, we present the following minimal reward function design.
\begin{itemize}
    \item \textbf{Success}: If the goal object $x$ specified by the instruction $I^{k,x}$ is reached correctly and the required interaction $k$ is performed, the agent is judged to be successful and receives a success reward.
    \item \textbf{Timeout}: Timeout penalty is given when the maximum number of steps $T$ is reached.
    \item \textbf{Failure}: If the agent reaches an incorrect goal or executes incorrect interactions, it is given a failure reward. At this juncture, the agent is not aware of which specific goal was incorrectly reached or which erroneous interaction was performed.
    \item \textbf{Timestep Penalty}: The agent receives a penalty reward for every step to accelerate trial and error.
\end{itemize}





\section{Method}
\label{sec:method}

\begin{algorithm}[t]
\caption{VHS with A3C}
\label{alg:vhs}
\begin{algorithmic}
\STATE Initialize actor and critic $\pi$ and $V$
\STATE Initialize representation parameters: $\theta$ and $\phi$ 
\STATE Goal Storage: $\mathcal{D}_g \leftarrow \emptyset$
\STATE Global, thread step counter : $T \leftarrow 0$, $t \leftarrow 0$
\REPEAT
    \STATE $t \leftarrow 0$ \COMMENT{Environment Reset}
    \STATE Hindsight Episode Buffer: $\mathcal{D}_f \leftarrow \emptyset$
    \STATE Get observation $o_0$, instruction $I^{k,x}$ from env 
    \STATE Get Prototypical Goal embedding $p^{x}$ from eq.~\ref{eq:proto}
    \REPEAT
        \STATE $a_t \sim \pi (a_t|o_t,p^{x}, k;\theta)$
        \STATE Receive reward $r_t$ and next observation $o_{t+1}$
        \STATE Store the transition $(o_t, a_t, V_{\phi}(o_t| p^x, k))$ in $\mathcal{D}_f$
        \IF{Success}
            \STATE $\mathcal{D}_g \leftarrow \mathcal{D}_g \cup \{ (o_t, I^x) \} $
            \COMMENT{Collect success states}
        \ENDIF
        \STATE $t \leftarrow t+1$ 
        \STATE $T \leftarrow T+1$
    \UNTIL terminal $o_t$ \OR $t = t_{max}$

    \FOR{$i \in \{ t-1, ...,0 \}$} 
        \STATE Calculate $\mathcal{L}_{RL}$  with Eq.~\ref{eq:actor} and ~\ref{eq:critic} \\ 
    \ENDFOR \algorithmiccomment{Calculate A3C loss}
    
    \STATE Calculate $\mathcal{L}_{s}$ from $D_{g}$ with Eq.~\ref{eq:supcon}\algorithmiccomment{Goal-aware loss} 
    
    \IF{Fail \AND Random(0,1) $< \eta$}
        \STATE Re-labeling $p_{x}$ to $f(o_T)$ and $k$ to $a_T$ in $\mathcal{D}_f$
        \STATE Calculate $\mathcal{L}_{VHS}$ from $D_f$ with  Eq.~\ref{eq:vhs_loss}
    \ENDIF
    \algorithmiccomment{Calculate VHS loss}
    \STATE Update the parameters $\theta$ and $\phi$ using loss $\mathcal{L}$ in Eq.~\ref{eq:total}  
\UNTIL $T > T_{max}$
\end{algorithmic}
\end{algorithm}

\subsection{Instruction to Prototypical Goal Embedding}
\label{sec:i2pg}
A crucial aspect of hindsight methods is their ability to redefine the goals of failed episodes.
However, in visual navigation tasks \cite{kim2021goal,kim2023sa,wu2018building,chaplot2020object,savva2019habitat} with multiple objects or targets, where goal is given by words or instructions, setting the terminal observation as a new goal poses a practical problem.
In general, the tasks specify various goals to be reached as representations within an embedding space, such as word embeddings.
Yet, when an agent fails to reach a desired goal, representing the final location of the failed episode in word form is not straightforward, in contrast to tasks where goals are given as coordinates.
For instance, if the agent is instructed to navigate to and pick up a ketchup but instead picks up a mustard, then re-labeling this terminal observation as ``Pick up a mustard'' is difficult without an additional external signal of which object the agent has arrived at.


To address this issue, we introduce the Prototypical Goal (PG) embedding method for instructions, as depicted in Figure~\ref{fig:overall}. This method diverges from the typical approach of word embedding instructions. Instead, it substitutes the instruction input with feature representations of the target $x$ with which the agent is supposed to interact. PG utilizes the goal observations stored during the learning process to replace conventional input instructions with prototypical representations.
These representations $p^x$ are computed as the average vector of the embedded goal observations associated with each class $x$ as defined by the following equation:
\begin{equation}
\label{eq:proto}
    p^x = \frac{1}{|D_g^{x}|} \sum_{o \in D_{g}^x} f(o).
\end{equation}
In the above equation, $f(\cdot) : \mathcal{O} \rightarrow \mathcal{Z}$ is a convolution neural network that extracts the features of the input observation, $\mathcal{Z}$ is an observation embedding space, and $D_g^x$ is $D_g$ with object $x$.
If there is no goal observation for $I^{k,x}$ in the agent's experience $D_g^x$, $p^x$ is instead a random vector sampled from $\mathcal N(\mathbf{0}, 1)$.

This method is similar to prototypical networks for few-shot learning \cite{snell2017prototypical,ji2020improved,pahde2021multimodal} and we use our embedding method to guide the pursuit of a prototype of the desired goal, akin to strategies employed in scene-driven navigation.





\subsection{Hindsight Episode Buffer}

To solve the challenging problem of sparse reward settings, HER learns by replacing unsuccessful experiences in the agent's experience with successful ones by re-labeling the goal.
These re-labeled experiences can serve as a textbook for exemplary behavior for the agent, and by exploiting it, the agent indirectly drives deep exploration.
We take this approach and apply it to interactive visual navigation that use re-labeling to change the agent's experience as shown in Figure~\ref{fig:overall}.
When the instruction is $I^{k,x}$, there are three cases of rewards, outlined in Sec.~\ref{sec:rwd_settings}.
Among these, we choose to perform re-labeling when failure reward is encountered, i.e., when incorrect goal is reached or incorrect interaction is performed.
In such cases, the observation and action at final time step $T$ can be readily used as the new goal and interaction.
Thus, upon receiving failure reward, the goal and interaction are re-labeled from $(p^x, k)$ to $(f(o_T), a_T)$.
Note that PG embedding from Sec.~\ref{sec:i2pg} precisely makes such re-labeling of goal possible, by placing both $f(o_T)$ and $p^x$ in the same embedding space $\mathcal{Z}$.
The experience converted by this process, while successful, is a suboptimal path because it is an unintended experience from a suboptimal policy.
We describe how to utilize this suboptimal path for learning in Sec.~\ref{sec:vhs}.
%


\begin{figure*}[t]
\label{fig:main}
  \centering
  \begin{subfigure}{0.32\linewidth}
    \includegraphics[width=\linewidth]{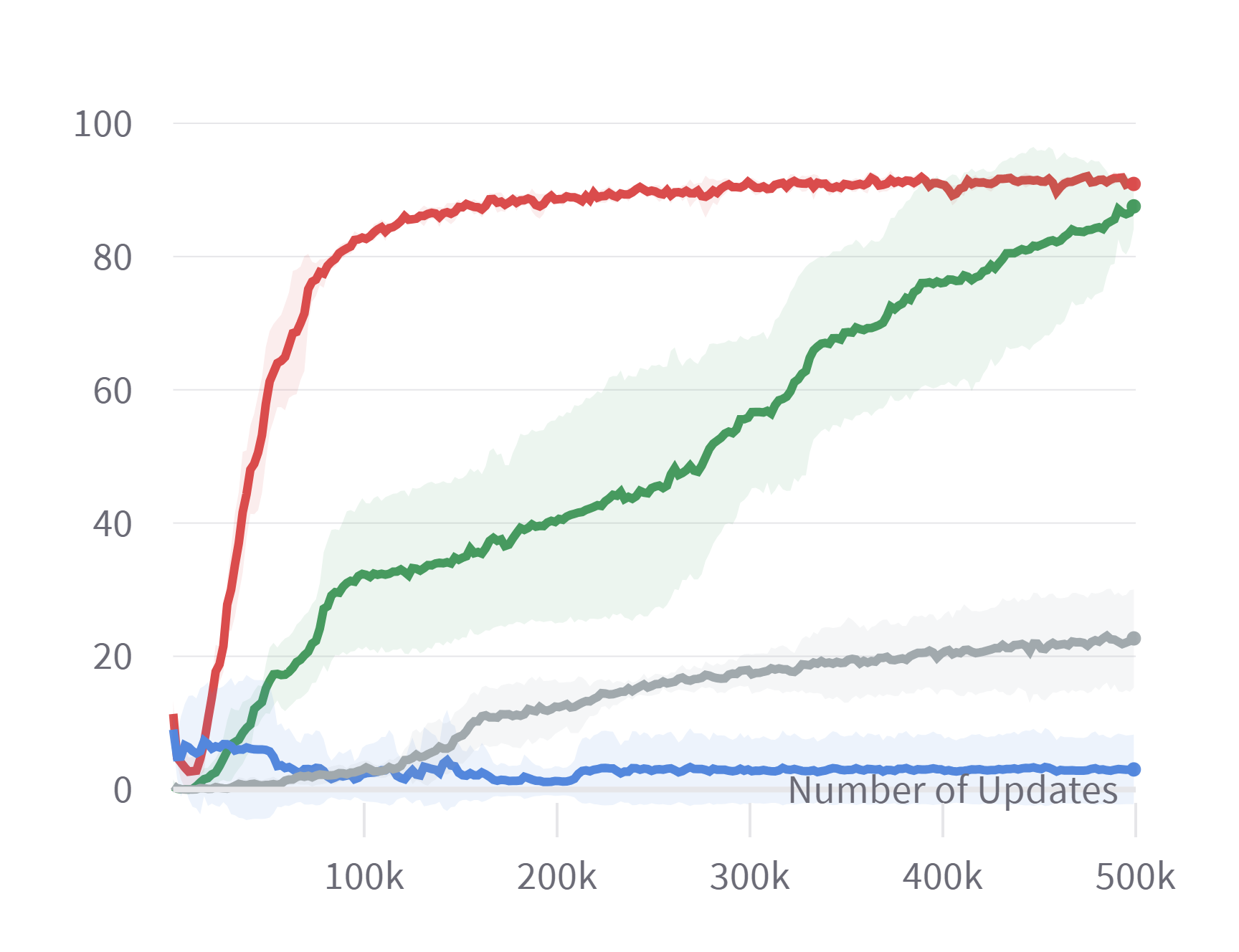}
    \caption{Task 1: Object Navigation}
    \label{fig:results_task1}
  \end{subfigure}
  \begin{subfigure}{0.32\linewidth}
    \includegraphics[width=\linewidth]{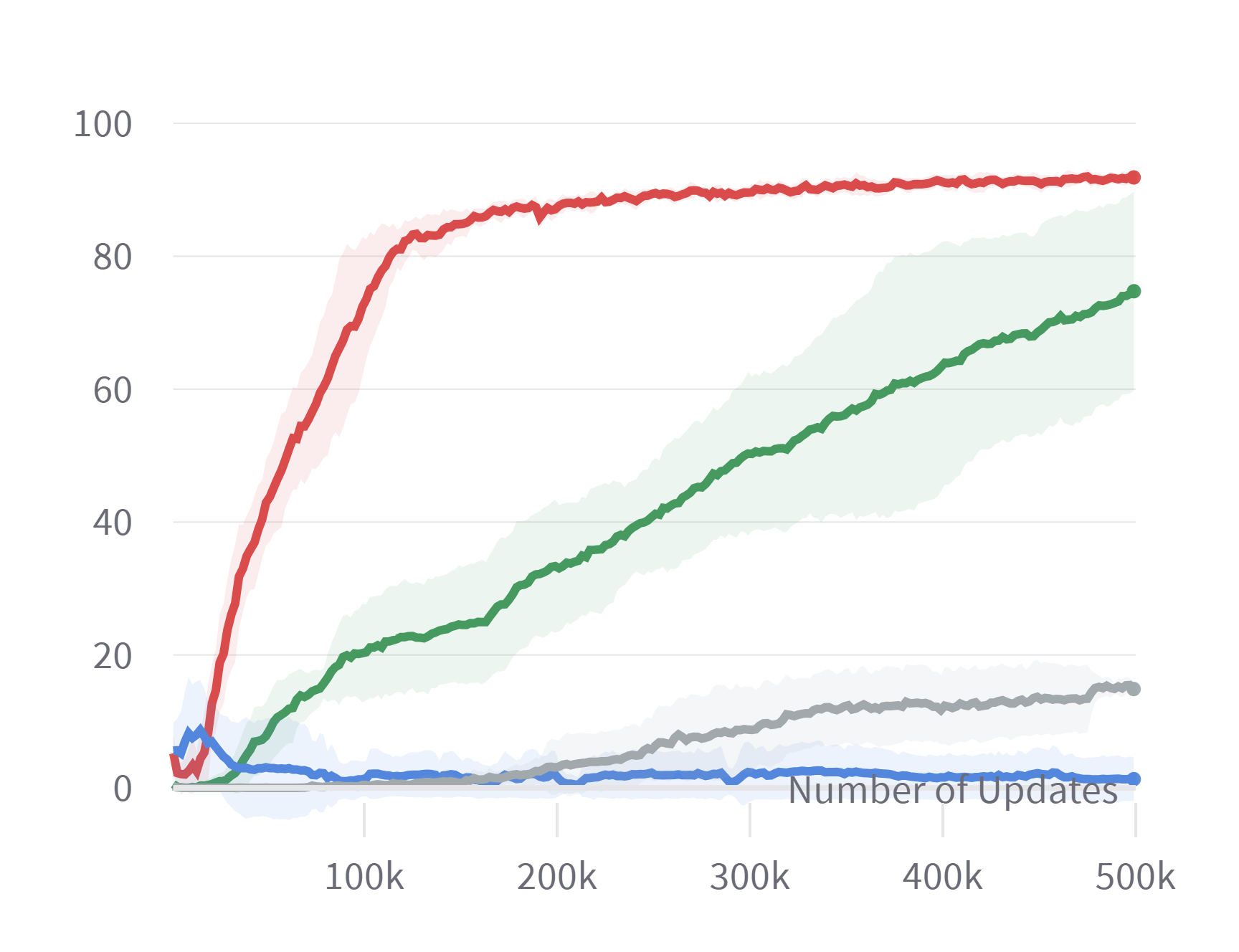}
    \caption{Task 2: Interactive Object Navigation}
    \label{fig:results_task2}
  \end{subfigure}
  \begin{subfigure}{0.32\linewidth}
    \includegraphics[width=\linewidth]{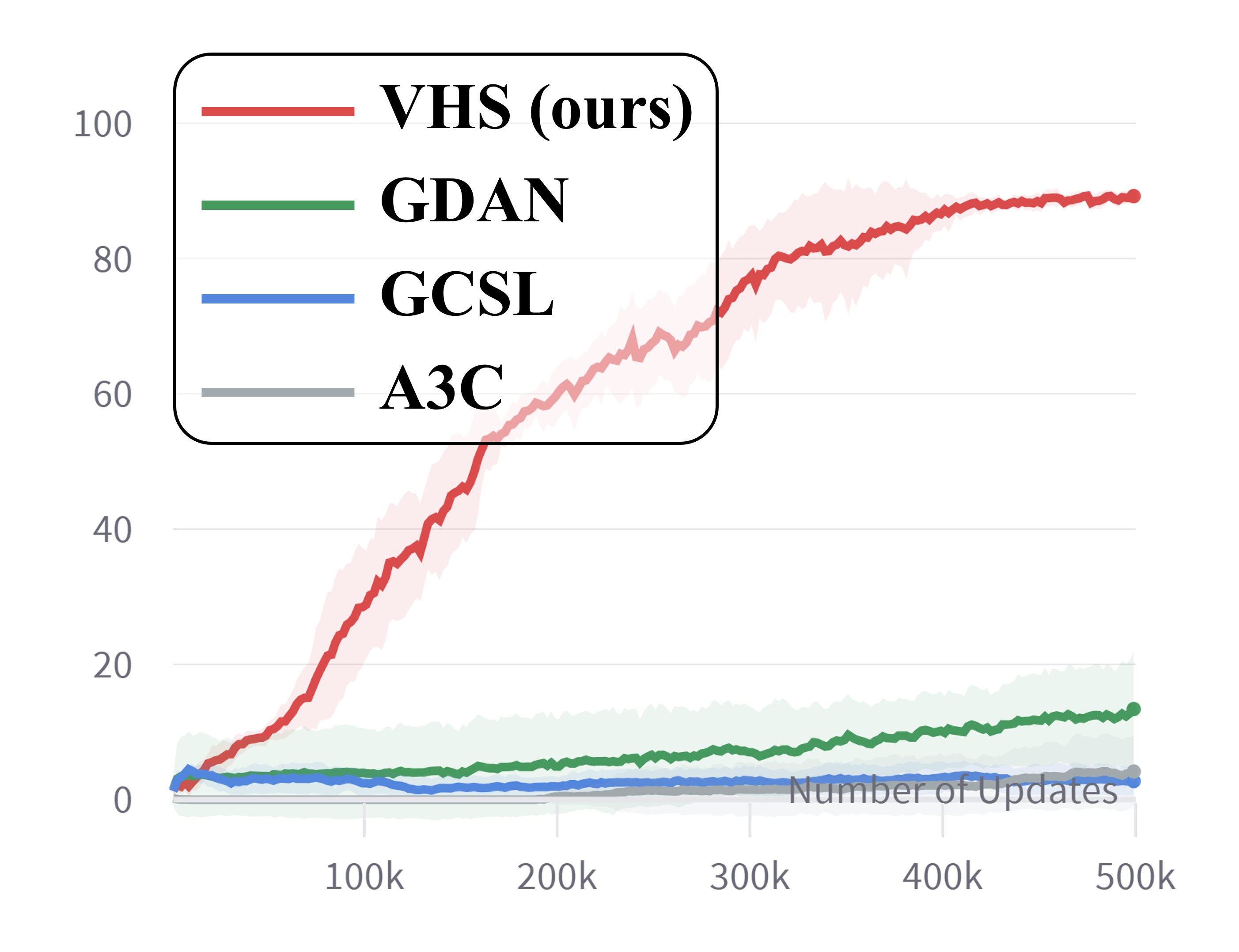}
    \caption{Task 3: Multi-interaction ObjNav}
    \label{fig:results_task3}
  \end{subfigure}
  \caption{Learning curves for three visual navigation tasks. In all tasks, our method shows rapid improvement and saturation in performance, demonstrating high sample efficiency. Especially in Task3, it shows a significant gap with baselines in task performance. The x-axis is number of updates and the y-axis is success rate. Each curve is produced from 7 trials and indicate bounds as mean $\pm$ standard deviation.}
  \label{fig:main}
\end{figure*}

\subsection{Visual Hindsight Self-Imitation Learning}
\label{sec:vhs}
We have previously explained an embedding method for re-labeling failed episodes in the vision domain in Sec.~\ref{sec:i2pg}.
Using this method, we propose self-imitation for online learning, using the newly obtained trajectories.
The overall algorithm is outlined in Algorithm~\ref{alg:vhs}.
First, for each episode, we collect the trajectory $\{(o_t, a_t, V(o_t|p^x, k))\}_{t=1}^{T}$ in a buffer $D_{f}$.
While calculating the RL loss for each episode, if it is a failed episode, we additionally compute the VHS loss according to the reward of the episode as shown below.
\begin{align}
    \mathcal{L}_{VHS} =& -\mathbb{E}_{o_i, a_i, v_i \in D_f} \log\pi(a_i|o_i,f(o_T),a_T)\label{eq:vhs_loss}\\
    &+ \lVert V(o_i | f(o_T), a_T) - v_i \rVert^2 \nonumber
\end{align}
Note that the data collected by the current agent is not accumulated after the agent learns through behavioral cloning of actions and values, because it is a suboptimal path. 
In this way, online learning with hindsight becomes possible.

Instructions of the failed episodes are re-labeled using PG embedding, and the observations and action and value of the episodes are imitated to learn.
Especially in hard exploration problems, learning is slow or unsuccessful because it is very difficult to obtain successful experiences. 
By generating informative experiences through the hindsight method and self-imitating them, the agent can solve difficult problems efficiently and effectively, even in the visual domain. 


Finally, the overall loss function looks like this:
\begin{equation}
    \label{eq:total}
    \mathcal{L} = \mathcal{L}_{RL} + \alpha\mathcal{L}_{s} + \beta \mathcal{L}_{VHS}
\end{equation}
We also use hyperparameter $\eta \in [0,1]$, where $\mathcal{L}_{VHS}$ is ignored in Eq.~\ref{eq:total} by probability $(1-\eta)$, in order to mitigate the agent overfitting to suboptimal re-labeled trajectory.

\section{Experiments}
\label{sec:exp}


\subsection{Experiments Setup}
Through experiments, we aim to evaluate whether our method can efficiently learn interactive visual navigation tasks with sparse reward settings.
In the experiments, we set up a fetch mobile robot in the MuJoCo environment \cite{todorov2012mujoco} where five objects (Mustard, BBQSauce, SaladDressing, OrangeJuice, Milk), which are 3D models from Hope3D dataset \cite{tyree2022hope}, are placed at random locations in each episode. 
At each time step, the agent can choose a movement action from $\mathcal{M} = \{MoveForward,TurnLeft,TurnRight\}$ or an interaction action from $\mathcal{K}$, where $\mathcal{K}$ depends on the given task as outlined in Sec.~\ref{sec:task_details}. 
In case the agent is not at the boundary of any object, an interaction action is neglected, treated as a no-op action.
An episode ends upon success/timeout/failure, and agent parameter updates are performed at the end of each episode. 
The evaluation measures the success rate with 300 episodes.
We set up three tasks in Sec.~\ref{sec:task_details} to evaluate that our method learns efficiently and effectively with various interaction settings.
Further details of the experimental details are described in the Appendix ~\ref{app:exp}.

We will make the environment used for our experiments and the source code of our proposed method publicly accessible upon the publication of our work.



\subsubsection{Task Details}
\label{sec:task_details}
In our tasks, the agent is located in the center of the room at episode reset.
The three tasks are as follows, in order of increasing difficulty.

\begin{itemize}
    \item \textbf{Task1} - \textbf{Object Navigation}: Success and failure are determined automatically when the agent finds and reaches within a certain boundary of each object. No interaction is required, i.e. $\mathcal{K} = \emptyset$. If the object specified by the instruction is reached, the episode is a success; otherwise, it is a failure.
    \item \textbf{Task2} - \textbf{Interactive Object Navigation}: Task2 is similar to Task1, except that the agent must perform a single interaction $\mathcal{K}=\{$\textit{Interaction}$\}$ with each object to determine success and failure.
    If an object boundary is reached but no interaction is performed, the episode continues.
    \item \textbf{Task3} - \textbf{Multi-interaction ObjNav}: In Task3, the number of interactions increases to three as  $\mathcal{K}=\{$\textit{Interaction1}, \textit{Interaction2}, \textit{Interaction3}$\}$, so the total number of instruction types increases to 15 with five objects. The instruction is given by environment randomly and the episode is considered a success if both the object and the interaction are correct, and a failure if either the object or the interaction is incorrect.
\end{itemize}


\begin{figure*}[t]
  \centering
  \begin{subfigure}{0.32\linewidth}
    \includegraphics[width=\linewidth]{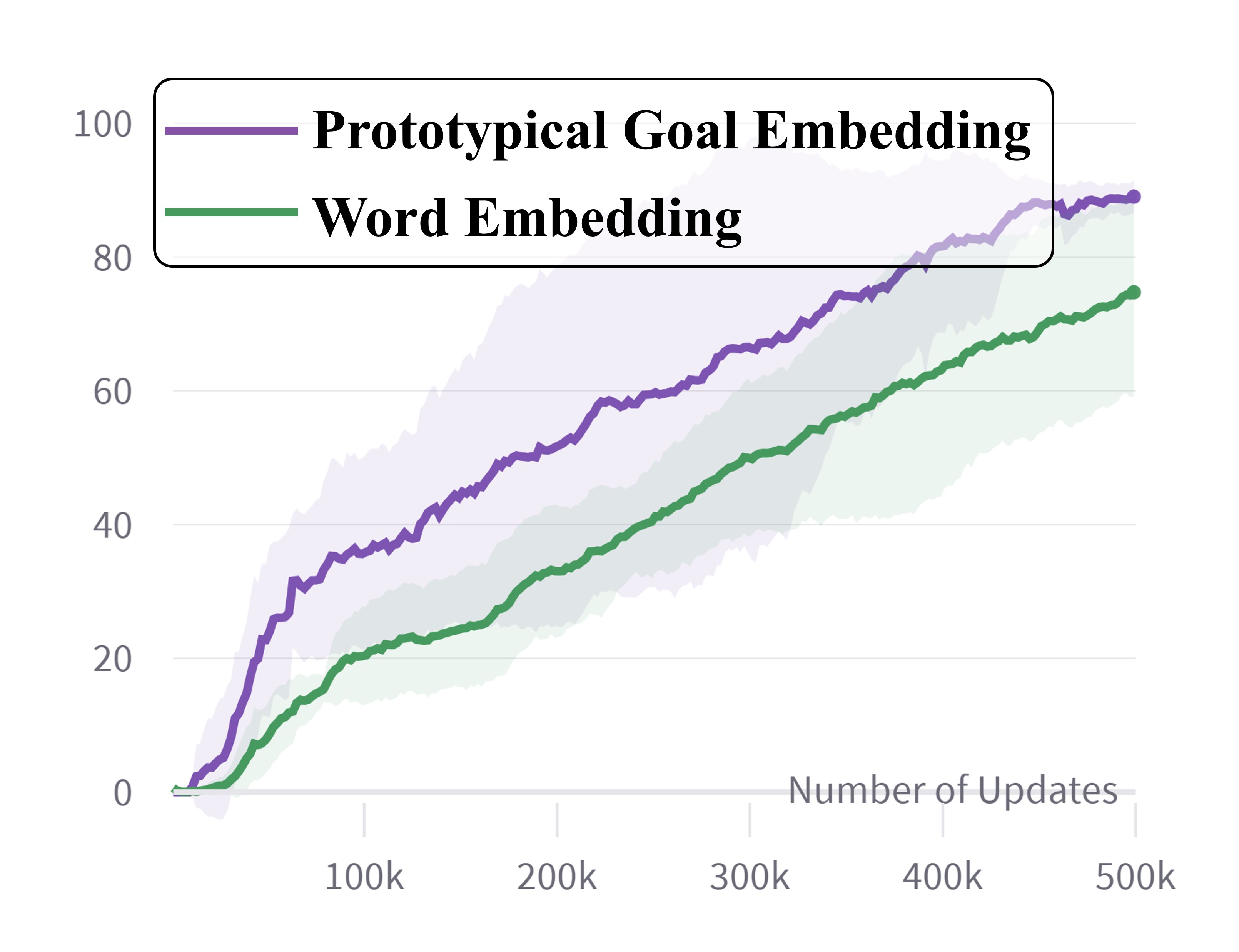}
    \caption{Ablation for Embedding methods}
    \label{fig:ablation_embedding}
  \end{subfigure}
  \begin{subfigure}{0.32\linewidth}
    \includegraphics[width=\linewidth]{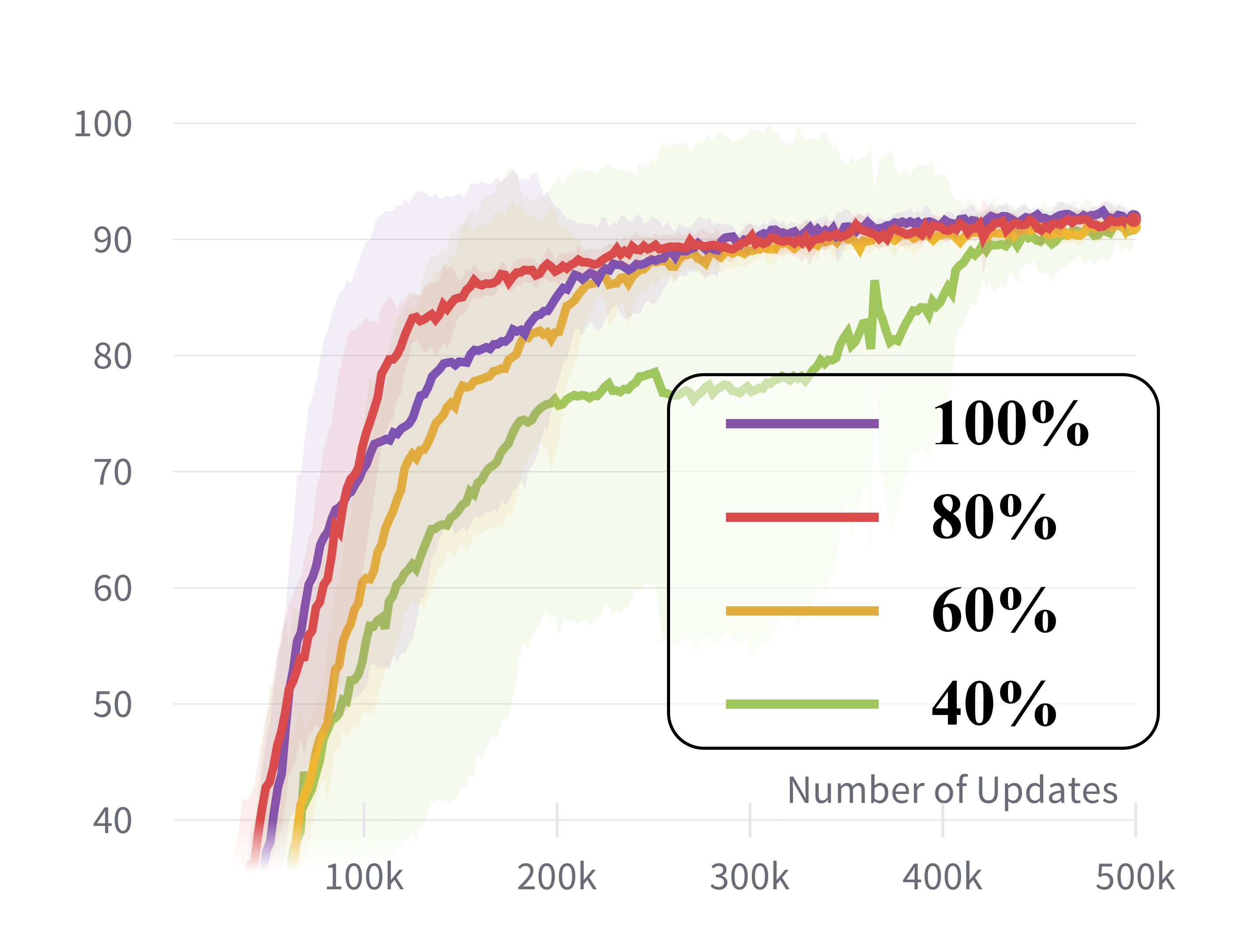}
    \caption{Ablation for triggering VHS}
    \label{fig:ablation_trigger}
  \end{subfigure}
  \begin{subfigure}{0.32\linewidth}
    \includegraphics[width=\linewidth]{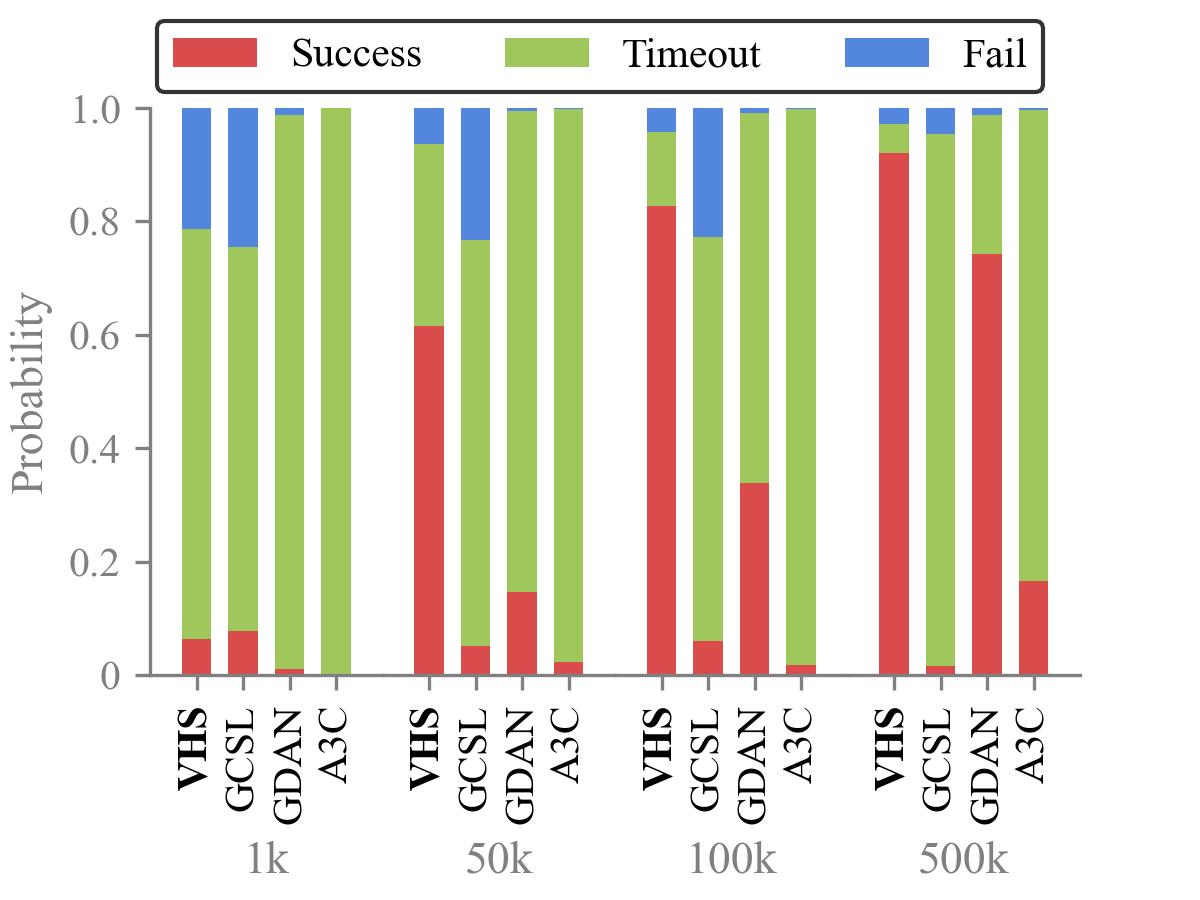}
    \caption{Proportion of three rewards types}
    \label{fig:ablation_reward}
  \end{subfigure}
  \caption{Learning curves of two ablation studies and an analysis of proportion of reward types. Experiments are performed on Task2 (Interactive Object Navigation), x-axis is number of updates, and y-axis is success rate in (a), (b), and reward proportion in (c).}
  \label{fig:ablation}
\end{figure*}



\subsubsection{Baselines}
We compare our method to the competitive methods below:
\begin{itemize}
    \item \textbf{A3C} \cite{mnih2016asynchronous}: The most basic reinforcement learning algorithm, it learns asynchronously in multiple environments through multiprocessing to accelerate learning speed. 
    \item \textbf{GDAN} (Goal-Discriminative Attention Networks) \cite{kim2021goal}:  It learns through cross-entropy loss using goals collected from visual navigation tasks for goal-aware learning. It proposes an agent that pursues goal-directed behavior, learns sample-efficiently, and further proposes an attention network to maximize the efficiency of the proposed method.  The instruction is input by word embedding method.
    \item \textbf{GCSL} (Goal-Conditioned Self-Supervised Learning) \cite{ghosh2020learning}: This method shows that learning is efficient using only self-imitation learning by re-labeling the next state or the end of the episode as the goal. Because this method does not account for visual navigation tasks, we add our proposed PG embedding to re-label the end of the episode as the goal, and also apply SupCon loss to learn about the goals. The main difference between GCSL and our method is that it performs self-imitation learning without a reward-maximizing RL policy.
    \item \textbf{VHS (Ours)} : In our proposed method, we calculate SupCon loss for learning the goals based on A3C. When learning, we use PG embeddings to pursue the goal, re-label the last observation and interaction in the failed episodes, and learn by self-imitation.
\end{itemize}
We iteratively trained with random seeds for at least 7 times, and the details of the hyperparameters used and additional ablations are shown in the Appendix ~\ref{app:add} and ~\ref{app:exp}.


\subsection{Results and Analyses}
Figure~\ref{fig:main} presents the learning trajectories of different models across three tasks, showcasing the efficacy of each approach. A3C demonstrates marginal progress, with a performance of $23.1 \pm 7.0 \%$ in Task1 and negligible improvement in Task3, suggesting that the three tasks are very challenging to learn with a conventional RL algorithm.

GDAN shows learning performance of $87.2 \pm 4.2 \%$ and $74.3 \pm 15.1 \%$ on the Task1 and Task2 in Figure~\ref{fig:results_task1} and Figure~\ref{fig:results_task2}, respectively, with slow and steady improvements in learning, indicating that an agent with goal-directed behavior can learn even under sparse reward designs.
However, in Figure~\ref{fig:results_task3} where GDAN achieves $13.3 \pm 8.3 \%$ in Task3, the learning curve improves very weakly, suggesting that the goal-directed behavior and attention models alone are not sufficient in this task, which requires a variety of interactions.

In contrast, GCSL underperforms compared to A3C in the first two tasks and fails to exhibit learning across all tasks. This outcome suggests that a combination of reinforcement learning policies and self-imitation learning through re-labeling without explicit rewards is inadequate for robust learning.

On the other hand, our VHS exhibits remarkable sample efficiency, achieving state-of-the-art success rates of $\mathbf{91.4} \pm \mathbf{0.8}\%$, $\mathbf{92.0} \pm \mathbf{0.8}\%$ and $\mathbf{89.5} \pm \mathbf{0.9} \%$ on the three tasks, respectively. Notably, in Task3, as depicted in Figure~\ref{fig:results_task3}, where successful trials are exceedingly sparse, our method reaches a saturation point in learning with a high success rate. These results demonstrate the effectiveness of the proposed approach that combines re-labeling with self-imitation learning to facilitate the acquisition of successful experiences in the visual navigation environments with sparse rewards.


\subsection{Ablation Studies}

\subsubsection{Embedding Methods}
To evaluate the efficacy of PG embedding in VHS, we contrast it with the conventional word embedding technique. 
We apply the word embedding method and our proposed prototypical goal embedding in Task2. 
The algorithm we used for training is based on SupCon loss $\mathcal{L}_{s}$ for goal-aware on A3C as $\mathcal{L}_{RL}$, applying each embedding method.
The learning curves are depicted in Figure~\ref{fig:ablation_embedding}. 
The learning curves reveal consistent progress for both the word embedding and our PG embedding methods. Notably, PG embedding facilitates more efficient learning. 
This indicates that, within our experimental framework, the agent better generates and follows prototypical goals using PG embedding—such as in tasks requiring scene-driven navigation—compared to traditional word embeddings.

\subsubsection{Probability of Triggering VHS Loss}

We calculate and update the VHS loss by re-labeling when the agent have failed episodes. 
The agent calculate VHS loss based on a certain probability, and perform an ablation study on the probability $\eta$ of four cases \{40\%, 60\%, 80\% and 100\%\}.
This ablation study is also performed in task2, and the learning curve for each probability of VHS loss is plotted in Figure~\ref{fig:ablation_trigger}. 
In addition to the 80\% probability used in the training, we performed experiments with four different probabilities, and concluded that there is no difference in the task success rate for all probability parameters.
However, there is a very close difference between 80\% and 100\%, with 80\% saturating slightly more reliably first, suggesting that the trajectories used for self-imitation learning are suboptimal paths and that some of the experience are noise or makes overfitting.
In conclusion, we can see from the figure that the learning curve improves and saturates faster as more trial of VHS are trained, indicating that our proposed method is effective and that higher probabilities contribute more to improving the success rate.

\begin{figure}[t]
  \centering
   \includegraphics[width=0.95\linewidth]{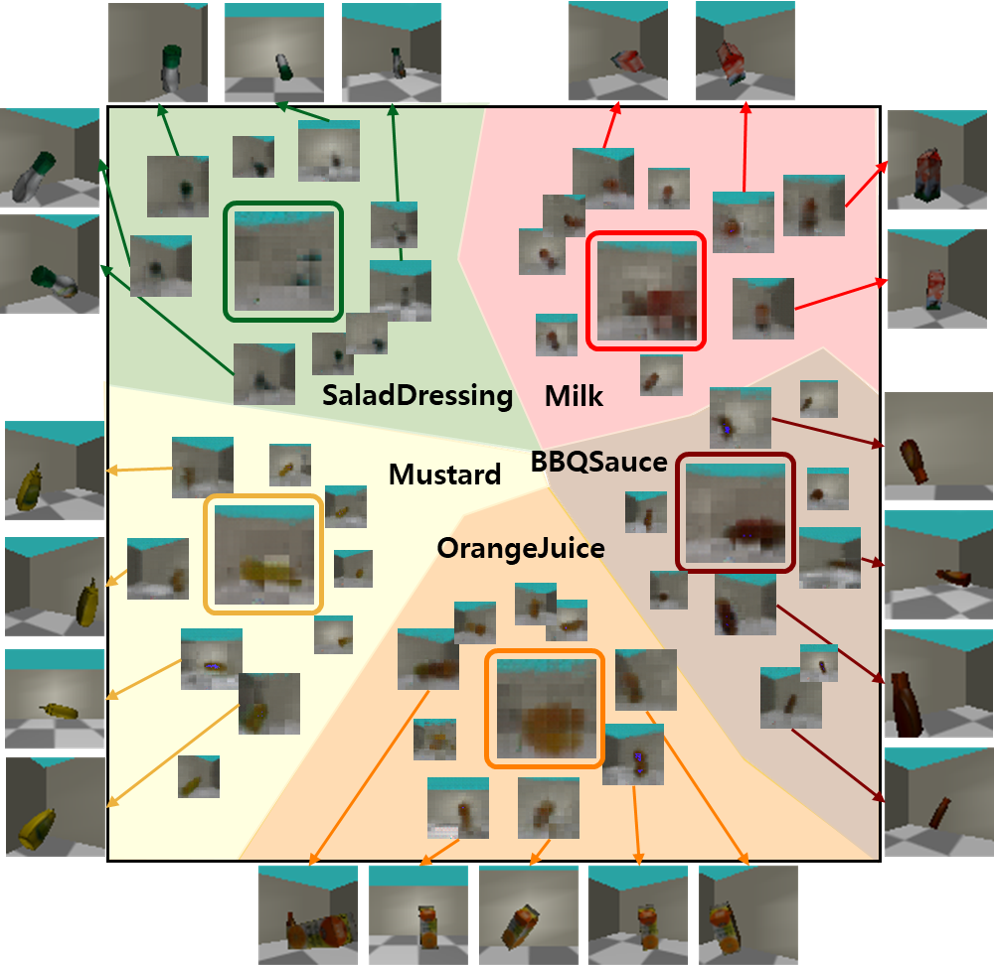}
   \caption{Visualization of prototypical goals and embeddings of goal observations. The center of each region shows the prototypical goal embedding of the corresponding object, while the neighbouring images visualize data from goal storage with close feature distances and point outward to the source images.}
   \label{fig:visualization}
\end{figure}


\subsubsection{Proportion of Reward Types}

Figure~\ref{fig:ablation_reward} shows the reward distribution across different methods during the learning process.
Timeouts are notably more frequent than successes and failures, a trend attributed to the agent’s rarity in achieving goals through random behavior.
Timeouts incur a lesser penalty (-0.1) than fail (-1.0), which initially encourages agents to time out rather than fail when success is less likely.

The A3C method consistently records the highest rate of timeouts and the lowest rate of fails, suggesting a tendency towards non-productive trial and error instead of successful task completion.
GCSL, despite occasionally reaching incorrect goals due to PG embedding and SupCon loss, shows prolonged periods of high fail rates. This method, which relies on imitation learning without RL incentives, suggests that failure to reach a goal does not significantly impact the learning outcomes.
GDAN predominantly encounters timeouts early in learning but exhibits gradual improvements in goal differentiation over time, leading to reduced timeouts. However, since GDAN does not account for interactions, its learning remains inefficient.
In contrast, VHS starts with a higher failure rate that diminishes over time. It enhances learning through self-imitation and penalization of unsuccessful attempts. Consequently, VHS achieves the highest success rates and sample efficiency.

\subsection{Visualization of Prototypical Goals}

We visualize the features extracted by prototypical goal embedding for the goal pursued by the agent.
To perform the visualization, we add a decoder training by Variational AutoEncoder (VAE) \cite{kingma2013auto,kingma2019introduction} with the proposed method, and train the decoder by sampling from the goal storage. Then, the prototypical goal embeddings obtained with Eq.~\ref{eq:proto} are visualized using the decoder.

The visualization of the five objects used in our experiments is shown in Figure~\ref{fig:visualization}, where we visualize the PG embeddings in the center of each region. The surrounding images are sampled from the goal storage with close feature distances and visualized in the same way, with the original image pointing to the outside.
Since each object is placed in different orientations and positions in the environment, we can see that the visualization of the embedding reflects the shape and color of objects.
The surrounding images also reflect the shape and color well.
For the goal pursued by the agent, it is shown that these visual aspect of the desired goal can be well-extracted and pursued through PG embedding. 
In addition, when re-labeling is performed in the fail episode, it is visually shown that PG embedding-like features can be set as the goal even for the incorrect goal similar to PG embedding.


\section{Conclusion}

In this paper, we present VHS approach, designed to enhance the learning performance in interactive visual navigation task with sparse rewards. 
Our comprehensive experimental results reveal that VHS significantly improves success rates and sample efficiency by hindsight visual goal re-labeling of unsuccessful episodes and by boosting deep exploration through leveraging self-imitation with enriched successful episodes. 
Notably, our PG embedding, crucial for enabling effective goal re-labeling from visual observations, operates effectively within this framework without reliance on any pre-trained models.
This research demonstrates the feasibility using the hindsight experience replay technique in real-time vision-based tasks, eliminating the need for prior data collection. 
It paves the way for the integration of this approach into a range of related fields. 
Moreover, we anticipate that subsequent research will validate the utility of VHS in incremental learning, especially for effectively interacting with new objects. 
A current limitation, however, is that the approach has not been evaluated in settings that require a continuous action space.



{
    \small
    \bibliographystyle{ieeenat_fullname}
    \bibliography{main}

\begin{thebibliography}{61}
\providecommand{\natexlab}[1]{#1}
\providecommand{\url}[1]{\texttt{#1}}
\expandafter\ifx\csname urlstyle\endcsname\relax
  \providecommand{\doi}[1]{doi: #1}\else
  \providecommand{\doi}{doi: \begingroup \urlstyle{rm}\Url}\fi

\bibitem[Al-Halah et~al.(2022)Al-Halah, Ramakrishnan, and Grauman]{al2022zero}
Ziad Al-Halah, Santhosh~Kumar Ramakrishnan, and Kristen Grauman.
\newblock Zero experience required: Plug \& play modular transfer learning for semantic visual navigation.
\newblock In \emph{Proceedings of the IEEE/CVF Conference on Computer Vision and Pattern Recognition}, pages 17031--17041, 2022.

\bibitem[Andrychowicz et~al.(2017)Andrychowicz, Wolski, Ray, Schneider, Fong, Welinder, McGrew, Tobin, Pieter~Abbeel, and Zaremba]{andrychowicz2017hindsight}
Marcin Andrychowicz, Filip Wolski, Alex Ray, Jonas Schneider, Rachel Fong, Peter Welinder, Bob McGrew, Josh Tobin, OpenAI Pieter~Abbeel, and Wojciech Zaremba.
\newblock Hindsight experience replay.
\newblock \emph{Advances in neural information processing systems}, 30, 2017.

\bibitem[Burda et~al.(2018)Burda, Edwards, Storkey, and Klimov]{burda2018exploration}
Yuri Burda, Harrison Edwards, Amos Storkey, and Oleg Klimov.
\newblock Exploration by random network distillation.
\newblock \emph{arXiv preprint arXiv:1810.12894}, 2018.

\bibitem[Chaplot et~al.(2018)Chaplot, Sathyendra, Pasumarthi, Rajagopal, and Salakhutdinov]{chaplot2018gated}
Devendra~Singh Chaplot, Kanthashree~Mysore Sathyendra, Rama~Kumar Pasumarthi, Dheeraj Rajagopal, and Ruslan Salakhutdinov.
\newblock Gated-attention architectures for task-oriented language grounding.
\newblock In \emph{Proceedings of the AAAI Conference on Artificial Intelligence}, 2018.

\bibitem[Chaplot et~al.(2020)Chaplot, Gandhi, Gupta, and Salakhutdinov]{chaplot2020object}
Devendra~Singh Chaplot, Dhiraj~Prakashchand Gandhi, Abhinav Gupta, and Russ~R Salakhutdinov.
\newblock Object goal navigation using goal-oriented semantic exploration.
\newblock \emph{Advances in Neural Information Processing Systems}, 33:\penalty0 4247--4258, 2020.

\bibitem[Dai et~al.(2020)Dai, Liu, and Anthony~Bharath]{dai2020episodic}
Tianhong Dai, Hengyan Liu, and Anil Anthony~Bharath.
\newblock Episodic self-imitation learning with hindsight.
\newblock \emph{Electronics}, 9\penalty0 (10):\penalty0 1742, 2020.

\bibitem[Dai et~al.(2021)Dai, Liu, Arulkumaran, Ren, and Bharath]{dai2021diversity}
Tianhong Dai, Hengyan Liu, Kai Arulkumaran, Guangyu Ren, and Anil~Anthony Bharath.
\newblock Diversity-based trajectory and goal selection with hindsight experience replay.
\newblock In \emph{PRICAI 2021: Trends in Artificial Intelligence: 18th Pacific Rim International Conference on Artificial Intelligence, PRICAI 2021, Hanoi, Vietnam, November 8--12, 2021, Proceedings, Part III 18}, pages 32--45. Springer, 2021.

\bibitem[Das et~al.(2018)Das, Datta, Gkioxari, Lee, Parikh, and Batra]{das2018embodied}
Abhishek Das, Samyak Datta, Georgia Gkioxari, Stefan Lee, Devi Parikh, and Dhruv Batra.
\newblock Embodied question answering.
\newblock In \emph{Proceedings of the IEEE conference on computer vision and pattern recognition}, pages 1--10, 2018.

\bibitem[Devo et~al.(2020)Devo, Mezzetti, Costante, Fravolini, and Valigi]{devo2020towards}
Alessandro Devo, Giacomo Mezzetti, Gabriele Costante, Mario~L Fravolini, and Paolo Valigi.
\newblock Towards generalization in target-driven visual navigation by using deep reinforcement learning.
\newblock \emph{IEEE Transactions on Robotics}, 36\penalty0 (5):\penalty0 1546--1561, 2020.

\bibitem[Duan et~al.(2022)Duan, Yu, Tan, Zhu, and Tan]{duan2022survey}
Jiafei Duan, Samson Yu, Hui~Li Tan, Hongyuan Zhu, and Cheston Tan.
\newblock A survey of embodied ai: From simulators to research tasks.
\newblock \emph{IEEE Transactions on Emerging Topics in Computational Intelligence}, 6\penalty0 (2):\penalty0 230--244, 2022.

\bibitem[Ecoffet et~al.(2019)Ecoffet, Huizinga, Lehman, Stanley, and Clune]{ecoffet2019go}
Adrien Ecoffet, Joost Huizinga, Joel Lehman, Kenneth~O Stanley, and Jeff Clune.
\newblock Go-explore: a new approach for hard-exploration problems.
\newblock \emph{arXiv preprint arXiv:1901.10995}, 2019.

\bibitem[Fang et~al.(2020)Fang, Zhu, Savarese, and Fei-Fei]{fang2020adaptive}
Kuan Fang, Yuke Zhu, Silvio Savarese, and Li Fei-Fei.
\newblock Adaptive procedural task generation for hard-exploration problems.
\newblock \emph{arXiv preprint arXiv:2007.00350}, 2020.

\bibitem[Fang et~al.(2018)Fang, Zhou, Shi, Gong, Xu, and Zhang]{fang2018dher}
Meng Fang, Cheng Zhou, Bei Shi, Boqing Gong, Jia Xu, and Tong Zhang.
\newblock Dher: Hindsight experience replay for dynamic goals.
\newblock In \emph{International Conference on Learning Representations}, 2018.

\bibitem[Fang et~al.(2019)Fang, Zhou, Du, Han, and Zhang]{fang2019curriculum}
Meng Fang, Tianyi Zhou, Yali Du, Lei Han, and Zhengyou Zhang.
\newblock Curriculum-guided hindsight experience replay.
\newblock \emph{Advances in neural information processing systems}, 32, 2019.

\bibitem[Fang et~al.(2022)Fang, Xu, Wang, and Zeng]{fang2022target}
Qiang Fang, Xin Xu, Xitong Wang, and Yujun Zeng.
\newblock Target-driven visual navigation in indoor scenes using reinforcement learning and imitation learning.
\newblock \emph{CAAI Transactions on Intelligence Technology}, 7\penalty0 (2):\penalty0 167--176, 2022.

\bibitem[Franklin(1997)]{franklin1997autonomous}
Stan Franklin.
\newblock Autonomous agents as embodied ai.
\newblock \emph{Cybernetics \& Systems}, 28\penalty0 (6):\penalty0 499--520, 1997.

\bibitem[Ghosh et~al.(2020)Ghosh, Gupta, Reddy, Fu, Devin, Eysenbach, and Levine]{ghosh2020learning}
Dibya Ghosh, Abhishek Gupta, Ashwin Reddy, Justin Fu, Coline~Manon Devin, Benjamin Eysenbach, and Sergey Levine.
\newblock Learning to reach goals via iterated supervised learning.
\newblock In \emph{International Conference on Learning Representations}, 2020.

\bibitem[Ho and Ermon(2016)]{ho2016generative}
Jonathan Ho and Stefano Ermon.
\newblock Generative adversarial imitation learning.
\newblock \emph{Advances in neural information processing systems}, 29, 2016.

\bibitem[Hussein et~al.(2017)Hussein, Gaber, Elyan, and Jayne]{hussein2017imitation}
Ahmed Hussein, Mohamed~Medhat Gaber, Eyad Elyan, and Chrisina Jayne.
\newblock Imitation learning: A survey of learning methods.
\newblock \emph{ACM Computing Surveys (CSUR)}, 50\penalty0 (2):\penalty0 1--35, 2017.

\bibitem[Ji et~al.(2020)Ji, Chai, Yu, Pang, and Zhang]{ji2020improved}
Zhong Ji, Xingliang Chai, Yunlong Yu, Yanwei Pang, and Zhongfei Zhang.
\newblock Improved prototypical networks for few-shot learning.
\newblock \emph{Pattern Recognition Letters}, 140:\penalty0 81--87, 2020.

\bibitem[Karnan et~al.(2022)Karnan, Warnell, Xiao, and Stone]{karnan2022voila}
Haresh Karnan, Garrett Warnell, Xuesu Xiao, and Peter Stone.
\newblock Voila: Visual-observation-only imitation learning for autonomous navigation.
\newblock In \emph{2022 International Conference on Robotics and Automation (ICRA)}, pages 2497--2503. IEEE, 2022.

\bibitem[Khosla et~al.(2020)Khosla, Teterwak, Wang, Sarna, Tian, Isola, Maschinot, Liu, and Krishnan]{khosla2020supervised}
Prannay Khosla, Piotr Teterwak, Chen Wang, Aaron Sarna, Yonglong Tian, Phillip Isola, Aaron Maschinot, Ce Liu, and Dilip Krishnan.
\newblock Supervised contrastive learning.
\newblock \emph{Advances in Neural Information Processing Systems}, 33:\penalty0 18661--18673, 2020.

\bibitem[Kim et~al.(2021)Kim, Lee, Kim, Ryu, Lee, and Zhang]{kim2021goal}
Kibeom Kim, Min~Whoo Lee, Yoonsung Kim, JeHwan Ryu, Minsu Lee, and Byoung-Tak Zhang.
\newblock Goal-aware cross-entropy for multi-target reinforcement learning.
\newblock \emph{Advances in Neural Information Processing Systems}, 34:\penalty0 2783--2795, 2021.

\bibitem[Kim et~al.(2023)Kim, Lee, Lee, Lee, Lee, and Zhang]{kim2023sa}
Kibeom Kim, Hyundo Lee, Min~Whoo Lee, Moonheon Lee, Minsu Lee, and Byoung-Tak Zhang.
\newblock L-sa: Learning under-explored targets in multi-target reinforcement learning.
\newblock \emph{arXiv preprint arXiv:2305.13741}, 2023.

\bibitem[Kingma and Welling(2013)]{kingma2013auto}
Diederik~P Kingma and Max Welling.
\newblock Auto-encoding variational bayes.
\newblock \emph{arXiv preprint arXiv:1312.6114}, 2013.

\bibitem[Kingma et~al.(2019)Kingma, Welling, et~al.]{kingma2019introduction}
Diederik~P Kingma, Max Welling, et~al.
\newblock An introduction to variational autoencoders.
\newblock \emph{Foundations and Trends{\textregistered} in Machine Learning}, 12\penalty0 (4):\penalty0 307--392, 2019.

\bibitem[Kolve et~al.(2017)Kolve, Mottaghi, Han, VanderBilt, Weihs, Herrasti, Gordon, Zhu, Gupta, and Farhadi]{ai2thor}
Eric Kolve, Roozbeh Mottaghi, Winson Han, Eli VanderBilt, Luca Weihs, Alvaro Herrasti, Daniel Gordon, Yuke Zhu, Abhinav Gupta, and Ali Farhadi.
\newblock {AI2-THOR: An Interactive 3D Environment for Visual AI}.
\newblock \emph{arXiv}, 2017.

\bibitem[Lee et~al.(2020)Lee, Lee, Hwang, and Zhang]{lee2020visual}
Chung-Yeon Lee, Hyundo Lee, Injune Hwang, and Byoung-Tak Zhang.
\newblock Visual perception framework for an intelligent mobile robot.
\newblock In \emph{2020 17th International Conference on Ubiquitous Robots (UR)}, pages 612--616. IEEE, 2020.

\bibitem[LI et~al.(2023)LI, Wang, and Tan]{li2023self}
Yao LI, YuHui Wang, and XiaoYang Tan.
\newblock Self-imitation guided high-efficient goal-conditioned reinforcement learning.
\newblock \emph{Available at SSRN 4419852}, 2023.

\bibitem[Luo et~al.(2021)Luo, Kasaei, and Schomaker]{luo2021self}
Sha Luo, Hamidreza Kasaei, and Lambert Schomaker.
\newblock Self-imitation learning by planning.
\newblock In \emph{2021 IEEE International Conference on Robotics and Automation (ICRA)}, pages 4823--4829. IEEE, 2021.

\bibitem[Lyu et~al.(2022)Lyu, Shi, and Zhang]{lyu2022improving}
Yunlian Lyu, Yimin Shi, and Xianggang Zhang.
\newblock Improving target-driven visual navigation with attention on 3d spatial relationships.
\newblock \emph{Neural Processing Letters}, 54\penalty0 (5):\penalty0 3979--3998, 2022.

\bibitem[Maksymets et~al.(2021)Maksymets, Cartillier, Gokaslan, Wijmans, Galuba, Lee, and Batra]{maksymets2021thda}
Oleksandr Maksymets, Vincent Cartillier, Aaron Gokaslan, Erik Wijmans, Wojciech Galuba, Stefan Lee, and Dhruv Batra.
\newblock Thda: Treasure hunt data augmentation for semantic navigation.
\newblock In \emph{Proceedings of the IEEE/CVF International Conference on Computer Vision}, pages 15374--15383, 2021.

\bibitem[Manela and Biess(2021)]{manela2021bias}
Binyamin Manela and Armin Biess.
\newblock Bias-reduced hindsight experience replay with virtual goal prioritization.
\newblock \emph{Neurocomputing}, 451:\penalty0 305--315, 2021.

\bibitem[Mezghan et~al.(2022)Mezghan, Sukhbaatar, Lavril, Maksymets, Batra, Bojanowski, and Alahari]{mezghan2022memory}
Lina Mezghan, Sainbayar Sukhbaatar, Thibaut Lavril, Oleksandr Maksymets, Dhruv Batra, Piotr Bojanowski, and Karteek Alahari.
\newblock Memory-augmented reinforcement learning for image-goal navigation.
\newblock In \emph{2022 IEEE/RSJ International Conference on Intelligent Robots and Systems (IROS)}, pages 3316--3323. IEEE, 2022.

\bibitem[Mnih et~al.(2016)Mnih, Badia, Mirza, Graves, Lillicrap, Harley, Silver, and Kavukcuoglu]{mnih2016asynchronous}
Volodymyr Mnih, Adria~Puigdomenech Badia, Mehdi Mirza, Alex Graves, Timothy Lillicrap, Tim Harley, David Silver, and Koray Kavukcuoglu.
\newblock Asynchronous methods for deep reinforcement learning.
\newblock In \emph{International conference on machine learning}, pages 1928--1937, 2016.

\bibitem[Oh et~al.(2018)Oh, Guo, Singh, and Lee]{oh2018self}
Junhyuk Oh, Yijie Guo, Satinder Singh, and Honglak Lee.
\newblock Self-imitation learning.
\newblock In \emph{International Conference on Machine Learning}, pages 3878--3887. PMLR, 2018.

\bibitem[Pahde et~al.(2021)Pahde, Puscas, Klein, and Nabi]{pahde2021multimodal}
Frederik Pahde, Mihai Puscas, Tassilo Klein, and Moin Nabi.
\newblock Multimodal prototypical networks for few-shot learning.
\newblock In \emph{Proceedings of the IEEE/CVF Winter Conference on Applications of Computer Vision}, pages 2644--2653, 2021.

\bibitem[Paine et~al.(2019)Paine, Gulcehre, Shahriari, Denil, Hoffman, Soyer, Tanburn, Kapturowski, Rabinowitz, Williams, et~al.]{paine2019making}
Tom~Le Paine, Caglar Gulcehre, Bobak Shahriari, Misha Denil, Matt Hoffman, Hubert Soyer, Richard Tanburn, Steven Kapturowski, Neil Rabinowitz, Duncan Williams, et~al.
\newblock Making efficient use of demonstrations to solve hard exploration problems.
\newblock \emph{arXiv preprint arXiv:1909.01387}, 2019.

\bibitem[Pshikhachev et~al.(2022)Pshikhachev, Ivanov, Egorov, and Shpilman]{pshikhachev2022self}
Georgiy Pshikhachev, Dmitry Ivanov, Vladimir Egorov, and Aleksei Shpilman.
\newblock Self-imitation learning from demonstrations.
\newblock \emph{arXiv preprint arXiv:2203.10905}, 2022.

\bibitem[Ramalingam et~al.(2020)Ramalingam, Yin, Rajesh~Elara, Tamilselvam, Mohan~Rayguru, Muthugala, and F{\'e}lix~G{\'o}mez]{ramalingam2020human}
Balakrishnan Ramalingam, Jia Yin, Mohan Rajesh~Elara, Yokhesh~Krishnasamy Tamilselvam, Madan Mohan~Rayguru, MA~Viraj~J Muthugala, and Braulio F{\'e}lix~G{\'o}mez.
\newblock A human support robot for the cleaning and maintenance of door handles using a deep-learning framework.
\newblock \emph{Sensors}, 20\penalty0 (12):\penalty0 3543, 2020.

\bibitem[Sahni et~al.(2019)Sahni, Buckley, Abbeel, and Kuzovkin]{sahni2019addressing}
Himanshu Sahni, Toby Buckley, Pieter Abbeel, and Ilya Kuzovkin.
\newblock Addressing sample complexity in visual tasks using her and hallucinatory gans.
\newblock \emph{Advances in Neural Information Processing Systems}, 32, 2019.

\bibitem[Savva et~al.(2019)Savva, Kadian, Maksymets, Zhao, Wijmans, Jain, Straub, Liu, Koltun, Malik, et~al.]{savva2019habitat}
Manolis Savva, Abhishek Kadian, Oleksandr Maksymets, Yili Zhao, Erik Wijmans, Bhavana Jain, Julian Straub, Jia Liu, Vladlen Koltun, Jitendra Malik, et~al.
\newblock Habitat: A platform for embodied ai research.
\newblock In \emph{Proceedings of the IEEE/CVF international conference on computer vision}, pages 9339--9347, 2019.

\bibitem[Schaal(1999)]{schaal1999imitation}
Stefan Schaal.
\newblock Is imitation learning the route to humanoid robots?
\newblock \emph{Trends in cognitive sciences}, 3\penalty0 (6):\penalty0 233--242, 1999.

\bibitem[Snell et~al.(2017)Snell, Swersky, and Zemel]{snell2017prototypical}
Jake Snell, Kevin Swersky, and Richard Zemel.
\newblock Prototypical networks for few-shot learning.
\newblock \emph{Advances in neural information processing systems}, 30, 2017.

\bibitem[Sun et~al.(2018)Sun, Bagnell, and Boots]{sun2018truncated}
Wen Sun, J~Andrew Bagnell, and Byron Boots.
\newblock Truncated horizon policy search: Combining reinforcement learning \& imitation learning.
\newblock \emph{arXiv preprint arXiv:1805.11240}, 2018.

\bibitem[Tang(2020)]{tang2020self}
Yunhao Tang.
\newblock Self-imitation learning via generalized lower bound q-learning.
\newblock \emph{Advances in neural information processing systems}, 33:\penalty0 13964--13975, 2020.

\bibitem[Tang and Kucukelbir(2021)]{tang2021hindsight}
Yunhao Tang and Alp Kucukelbir.
\newblock Hindsight expectation maximization for goal-conditioned reinforcement learning.
\newblock In \emph{International Conference on Artificial Intelligence and Statistics}, pages 2863--2871. PMLR, 2021.

\bibitem[Thakur et~al.(2023)Thakur, Sunbeam, Goecks, Novoseller, Bera, Lawhern, Gremillion, Valasek, and Waytowich]{thakur2023imitation}
Ravi~Kumar Thakur, MD-Nazmus~Samin Sunbeam, Vinicius~G. Goecks, Ellen Novoseller, Ritwik Bera, Vernon~J. Lawhern, Gregory~M. Gremillion, John Valasek, and Nicholas~R. Waytowich.
\newblock Imitation learning with human eye gaze via multi-objective prediction, 2023.

\bibitem[Todorov et~al.(2012)Todorov, Erez, and Tassa]{todorov2012mujoco}
Emanuel Todorov, Tom Erez, and Yuval Tassa.
\newblock Mujoco: A physics engine for model-based control.
\newblock In \emph{2012 IEEE/RSJ international conference on intelligent robots and systems}, pages 5026--5033. IEEE, 2012.

\bibitem[Tyree et~al.(2022)Tyree, Tremblay, To, Cheng, Mosier, Smith, and Birchfield]{tyree2022hope}
Stephen Tyree, Jonathan Tremblay, Thang To, Jia Cheng, Terry Mosier, Jeffrey Smith, and Stan Birchfield.
\newblock 6-dof pose estimation of household objects for robotic manipulation: An accessible dataset and benchmark.
\newblock In \emph{International Conference on Intelligent Robots and Systems (IROS)}, 2022.

\bibitem[Weihs et~al.(2021)Weihs, Deitke, Kembhavi, and Mottaghi]{RoomR}
Luca Weihs, Matt Deitke, Aniruddha Kembhavi, and Roozbeh Mottaghi.
\newblock Visual room rearrangement.
\newblock In \emph{IEEE/CVF Conference on Computer Vision and Pattern Recognition (CVPR)}, 2021.

\bibitem[Wijmans et~al.(2019)Wijmans, Kadian, Morcos, Lee, Essa, Parikh, Savva, and Batra]{wijmans2019dd}
Erik Wijmans, Abhishek Kadian, Ari Morcos, Stefan Lee, Irfan Essa, Devi Parikh, Manolis Savva, and Dhruv Batra.
\newblock Dd-ppo: Learning near-perfect pointgoal navigators from 2.5 billion frames.
\newblock In \emph{International Conference on Learning Representations}, 2019.

\bibitem[Wortsman et~al.(2019)Wortsman, Ehsani, Rastegari, Farhadi, and Mottaghi]{wortsman2019learning}
Mitchell Wortsman, Kiana Ehsani, Mohammad Rastegari, Ali Farhadi, and Roozbeh Mottaghi.
\newblock Learning to learn how to learn: Self-adaptive visual navigation using meta-learning.
\newblock In \emph{Proceedings of the IEEE/CVF conference on computer vision and pattern recognition}, pages 6750--6759, 2019.

\bibitem[Wu et~al.(2020)Wu, Gong, Xu, Manocha, Dong, and Wang]{wu2020towards}
Qiaoyun Wu, Xiaoxi Gong, Kai Xu, Dinesh Manocha, Jingxuan Dong, and Jun Wang.
\newblock Towards target-driven visual navigation in indoor scenes via generative imitation learning.
\newblock \emph{IEEE Robotics and Automation Letters}, 6\penalty0 (1):\penalty0 175--182, 2020.

\bibitem[Wu et~al.(2018)Wu, Wu, Gkioxari, and Tian]{wu2018building}
Yi Wu, Yuxin Wu, Georgia Gkioxari, and Yuandong Tian.
\newblock Building generalizable agents with a realistic and rich 3d environment.
\newblock \emph{arXiv preprint arXiv:1801.02209}, 2018.

\bibitem[Xia et~al.(2018)Xia, Zamir, He, Sax, Malik, and Savarese]{xia2018gibson}
Fei Xia, Amir~R Zamir, Zhiyang He, Alexander Sax, Jitendra Malik, and Silvio Savarese.
\newblock Gibson env: Real-world perception for embodied agents.
\newblock In \emph{Proceedings of the IEEE conference on computer vision and pattern recognition}, pages 9068--9079, 2018.

\bibitem[Xiang et~al.(2020)Xiang, Qin, Mo, Xia, Zhu, Liu, Liu, Jiang, Yuan, Wang, et~al.]{xiang2020sapien}
Fanbo Xiang, Yuzhe Qin, Kaichun Mo, Yikuan Xia, Hao Zhu, Fangchen Liu, Minghua Liu, Hanxiao Jiang, Yifu Yuan, He Wang, et~al.
\newblock Sapien: A simulated part-based interactive environment.
\newblock In \emph{Proceedings of the IEEE/CVF Conference on Computer Vision and Pattern Recognition}, pages 11097--11107, 2020.

\bibitem[Yi and Yi(2019)]{yi2019mobile}
Jae-Bong Yi and Seung-Joon Yi.
\newblock Mobile manipulation for the hsr intelligent home service robot.
\newblock In \emph{2019 16th International Conference on Ubiquitous Robots (UR)}, pages 169--173. IEEE, 2019.

\bibitem[Zeng et~al.(2021)Zeng, Weihs, Farhadi, and Mottaghi]{zeng2021pushing}
Kuo-Hao Zeng, Luca Weihs, Ali Farhadi, and Roozbeh Mottaghi.
\newblock Pushing it out of the way: Interactive visual navigation.
\newblock In \emph{Proceedings of the IEEE/CVF Conference on Computer Vision and Pattern Recognition}, pages 9868--9877, 2021.

\bibitem[Zhu et~al.(2017)Zhu, Mottaghi, Kolve, Lim, Gupta, Fei-Fei, and Farhadi]{zhu2017target}
Yuke Zhu, Roozbeh Mottaghi, Eric Kolve, Joseph~J Lim, Abhinav Gupta, Li Fei-Fei, and Ali Farhadi.
\newblock Target-driven visual navigation in indoor scenes using deep reinforcement learning.
\newblock In \emph{2017 IEEE international conference on robotics and automation (ICRA)}, pages 3357--3364. IEEE, 2017.

\bibitem[Zhu et~al.(2020)Zhu, Lin, Dai, and Zhou]{zhu2020off}
Zhuangdi Zhu, Kaixiang Lin, Bo Dai, and Jiayu Zhou.
\newblock Off-policy imitation learning from observations.
\newblock \emph{Advances in Neural Information Processing Systems}, 33:\penalty0 12402--12413, 2020.

\end{thebibliography}
}

\clearpage
\setcounter{page}{1}
\maketitlesupplementary


\section{Additional Experiments}
\label{app:add}
\subsection{Ablation on Re-Labeling of Goals}
For VHS, we propose Prototypical Goal embedding method to re-label the end of episode as goal, only in the case of failed episodes.
For an ablation study on re-labeling methods, we perform additional experiments by re-labeling when episode ends in a failure, when the episode timeout occurs, and when either a failure or a timeout occurs (referred to as \textit{Both}).
These experiments are conducted in Task2.
In the experiments, for timeout, we re-label the end of episode as goal when the maximum number of steps $T=20$ is reached within an episode.

\begin{figure}[t]
  \centering
   \includegraphics[width=0.8\linewidth]{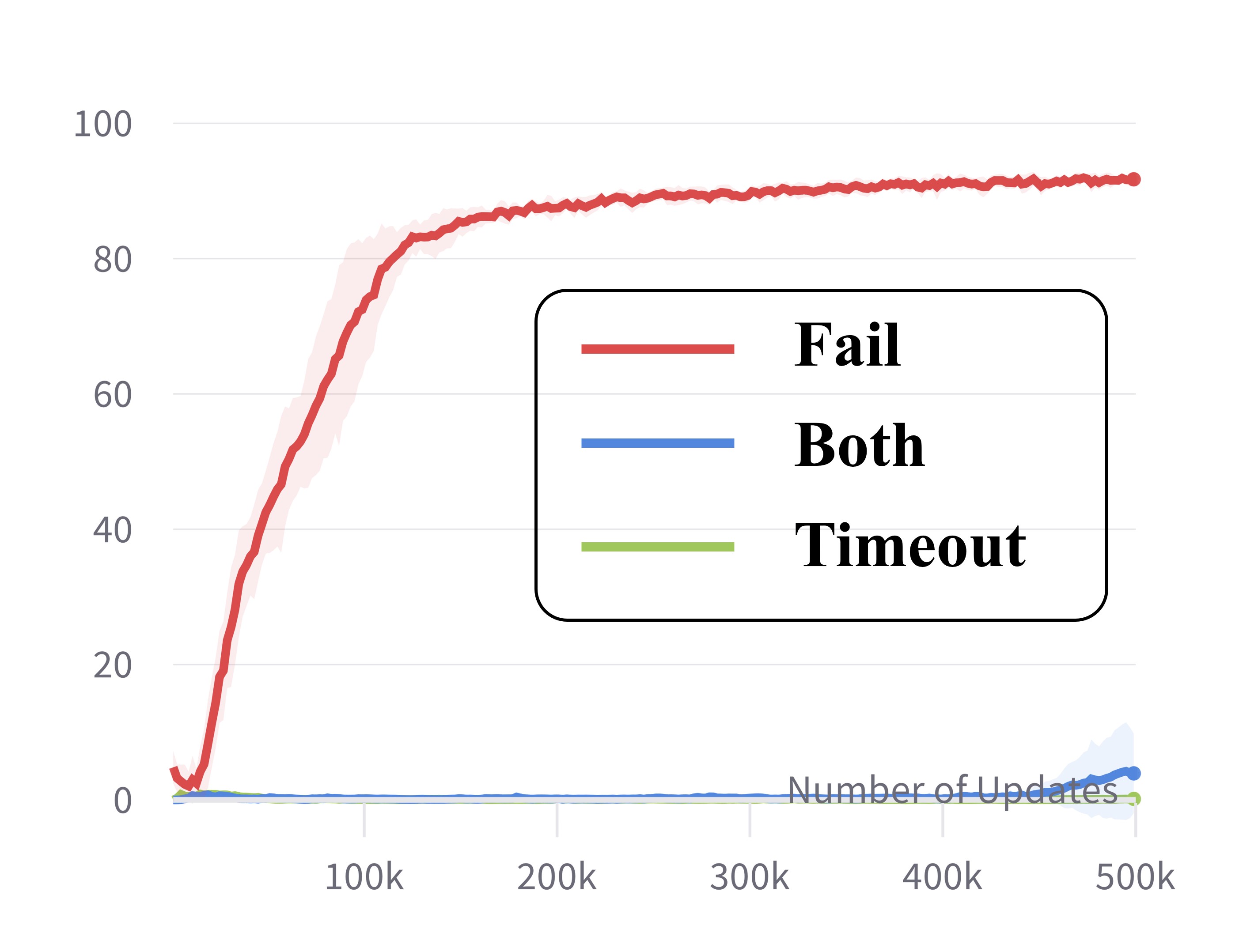}
   \caption{Ablation for types of re-labeling. Re-labeling on episode failure is more effective than re-labeling in other cases. }
    \label{fig:ablation_relabeling}
\end{figure}

Figure~\ref{fig:ablation_relabeling} shows the learning curves, and we can see that unlike the curve for re-labeling on failure, there is no improvement in performance when learning by re-labeling on timeout, and a slight improvement at the end for re-labeling on ``Both.''
When the agent ends the episode with timeout, the agent is still approaching the wall without sufficiently having learned to reach the goals, or it is performing unnecessary actions without reaching the goal that requires interaction.
For this reason, re-labeling on timeout is not considered appropriate.
Additionally, we can see a small improvement for re-labeling on ``Both,'' from which we speculate that re-labeling on timeout is mostly interfering with learning.
Finally, our results support that only the proposed re-labeling of failed episodes leads to efficient learning in an interactive visual navigation task.


\subsection{Visualization Using t-SNE}

\begin{figure}[t]
  \centering
   \includegraphics[width=0.9\linewidth]{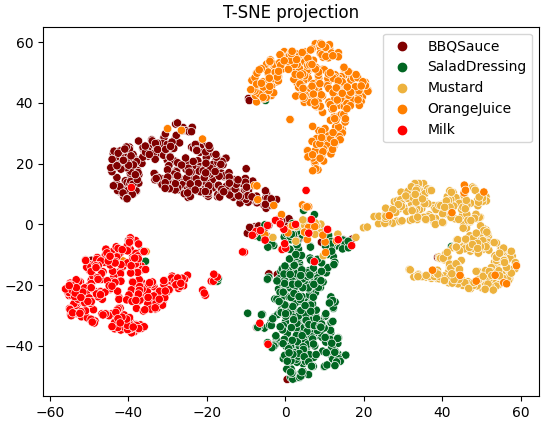}
   \caption{Visualization of goal observations using t-SNE. Note the distinct regions for each object, indicating that the features are well-segmented.}
    \label{fig:tsne}
\end{figure}

About 1,000 goal observations are collected for each class by an agent with uniform random policy, and the features are extracted by the learned VHS and visualized using t-SNE in Figure~\ref{fig:tsne}.
This figure shows that there are distinct regions for different classes, and the features for each goal are distinguishable once the representations are sufficiently trained.
This result supports Figure~\ref{fig:visualization} in the main text where the regions are well segmented by features.


\section{Experiments Details}
\label{app:exp}

\subsection{Experiments Details}

\subsubsection{Reward Settings}
\label{app:reward}
The reward settings for each task are as follows.
\begin{itemize}
    \item Success: If the goal is reached within a certain range, the episode is considered a success and the agent receives a reward of 10.0. For tasks that require interaction, if the agent does not perform the instructed interaction even after reaching the goal, it only receives a timestep penalty, without the episode terminating.
    \item Timeout: The reward for reaching the maximum number of steps $T$ in an episode is -0.1.
    \item Failure: If the agent reaches the wrong object or performs the wrong interaction, the episode is considered a failure. In this case, the agent receives a reward of -1.0.
    \item Timestep penalty: To encourage the agent to explore goals quicker, a reward of -0.01 is given to the agent every step.
\end{itemize}

\begin{table*}[t]
\centering
\begin{tabular}{l|rrr}
\toprule
Algorithm  & \textbf{Task1} (\%) & \textbf{Task2} (\%) & \textbf{Task3} (\%) \\
\midrule
A3C                         & 23.1 $\pm$ 7.0         & 15.5 $\pm$ 1.1   & 4.2 $\pm$ 5.7   \\
GCSL                        & 7.5 $\pm$ 8.9          & 8.7 $\pm$ 7.4  & 4.2 $\pm$ 0.9  \\
GDAN                        & 87.2 $\pm$ 4.2          & 74.3 $\pm$ 15.1   & 13.3 $\pm$ 8.3   \\
\textbf{VHS} (Ours)  & $\mathbf{91.4 \pm 0.8}$ & $\mathbf{92.0 \pm 0.8}$ & $\mathbf{89.5 \pm 0.9}$ \\
\bottomrule
Word Embedding (GDAN) & 87.2 $\pm$ 4.2 & 74.3 $\pm$ 15.1  & 13.3 $\pm$ 8.3 \\
\textbf{PG Embedding} (Ours)     & 90.6 $\pm$ 1.7    & 89.6 $\pm$ 2.1  &   31.8 $\pm$ 8.1 \\
\bottomrule
\end{tabular}
\caption{Success rate comparison for different algorithms and ablation study for different embedding methods on Task1, Task2, and Task3.}
\label{tab:task1}
\end{table*}

\begin{table}[t]
\centering
\begin{tabular}{l|rrr}
\toprule
Algorithm  & $\begin{matrix}\text{Number of}\\ \text{Updates} \end{matrix}$ & SRR (\%) $\downarrow$ & SEI (\%) $\uparrow$ \\
\midrule
GDAN        & 500,000          & 100   & 100   \\
\textbf{PG Embedding} & 345,116     &   69.0 & 222.8\\
\textbf{VHS} & 103,002 & $\mathbf{20.6}$ & $\mathbf{385.4}$ \\
\bottomrule
\end{tabular}
\caption{Sample efficiency measures in Task2. SRR (lower
the better) and SEI (higher the better) are measured with GDAN performance as a reference. “Number
of Updates” indicates the number of updates required to reach the reference performance.}
\label{tab:efficiency}
\end{table}

\subsubsection{Results Details}

Table~\ref{tab:task1} shows the performance of different baseline algorithms for Task1, Task2, and Task3, represented by the learning curve Figure~\ref{fig:main} in the main text.
The table also shows the results of the ablation study for different embedding methods.
For each setting, the best success rates for at least five runs are measured and averaged.
Our proposed method, VHS, shows the highest learning performance as well as the lowest variance in all tasks, converging stably in each iteration.
Particularly in Task3, VHS shows a significant gap compared to the baselines, indicating that our method are robust in sparse reward settings in interactive visual navigation tasks.
Furthermore, the ablation study for different embedding methods shows that the Prototypical Goal embedding method has higher performance and lower variance than GDAN with word embeddings, and experimentally supports that it can replace word embeddings.
The difference between the performance of PG embedding and VHS in all tasks is due to the fact that in the latter, self-imitation learning is performed after embedding through re-labeling.
This leads to a huge difference in terms of sample efficiency, especially in Figure~\ref{fig:results_task2}.


\subsubsection{Sample Efficiency Measure}

To quantitatively measure the sample efficiency, we use the Sample Requirement Ratio (SRR) and Sample Efficiency Improvement (SEI) metrics as proposed in \cite{kim2021goal}.
Taking algorithm $A$ as a reference, these measures are calculated based on the success rate $X$\% of this reference algorithm $A$ and the number of updates $n_A$ required for $A$  to reach the corresponding success rate.
Then we find the number of updates $n_B$ required to reach the success rate $X$\% for the algorithm $B$ we want to measure, and calculate $SRR = n_B/n_A$. The SEI is calculated as $(n_A-n_B)/n_B$. Lower SRR and higher SEI indicate better sample efficiency relative to the baseline. 

We show the measurements of the sample efficiency in Table~\ref{tab:efficiency}. We choose GDAN as the reference algorithm and take 74.3\% after 500k updates as the criterion in Task2 to measure the efficiency of PG embedding and VHS.
As shown in the table, the metrics show that the VHS method is significantly more sample-efficient than GDAN, with \textbf{SRR} of up to $\mathbf{20.6\%}$ and \textbf{SEI} of $\mathbf{385.4\%}$.
In addition, PG embedding has SRR of 69\% and SEI of 222.8\%, indicating that it also improves sample efficiency.



\subsubsection{Implementation Details}

We share the implementation details such as neural network architecture and hyperparameters used in the experiments.

The agent receives a 2-frame-stack of 84$\times$84 RGB images as an input.
The feature extractor processes this input and outputs 256 hidden features.
The feature extractor is a 4-layered Convolutional Neural Network, and batch normalization is applied to each layer.
All convolutional layers have kernel size 3, stride 2 and padding 1, and the output dimension is 256.
Afterwards, the input image and the PG embedding is processed via gated-attention \cite{chaplot2018gated}, and the resulting feature vector is fed into a Long Short-Term Memory (LSTM) module, where the output hidden vector and context vector have dimension of 256.
In Task3, which involves various interactions, the interactions given by the instruction are embedded and concatenated into the features before and after the LSTM.
Finally, the action and value are output by feeding the LSTM's output hidden vector into the policy and value function, each composed of a 2-layered Multi-Layer Perceptron (MLP).


\begin{figure}[t]
  \centering
   \includegraphics[width=0.8\linewidth]{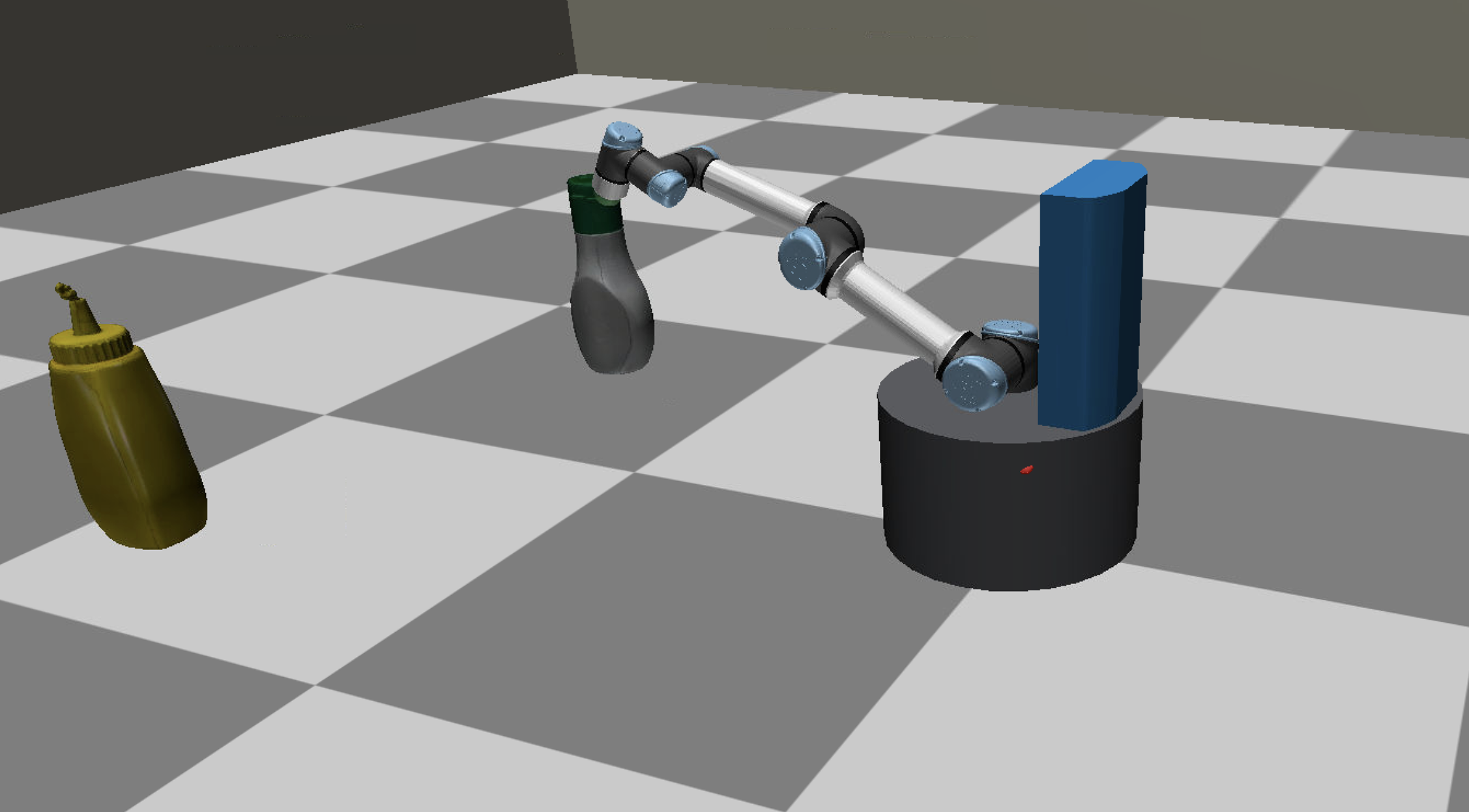}
   \caption{Example scene of our environment. At each reset, the agent is located at the center of the map, and the objects are placed in a randomized pose. We set up an agent to perform interactive visual navigation by combining a UR5 arm robot and a Fetch mobile robot.}
    \label{fig:env_sample}
\end{figure}


\begin{figure}[t]
  \centering
   \includegraphics[width=0.8\linewidth]{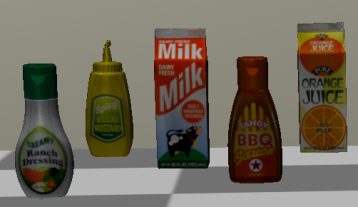}
   \caption{Images of the objects used in the experiment. From left to right, the objects are SaladDressing, Mustard, Milk, BBQSauce and OrangeJuice.}
    \label{fig:all_obj}
\end{figure}

\subsection{Environment Details}


The first-person perspective navigation environment we used for our experiments is shown in Figure~\ref{fig:env_sample} and the objects are shown in Figure~\ref{fig:all_obj}.
The map we used for our experiments is 3 m $\times$ 3 m in size, and the agent has a stride length of 0.5 meters.
At each episode reset, the agent is located in the center of the map, and the objects are located in random positions within the map boundaries and in random postures.
Our agent is a combination of a fetch mobile robot and a UR5 arm and is set up to perform the task of approaching and interacting with objects.
Our environment is characterized by the need to perform the task in a sparse reward setting, without a pre-trained model.
The five objects that need to be reached are illustrated in Figure~\ref{fig:all_obj}, all of which are placed in random postures, including rotation, in the experiments. 
All of the objects are easily encountered in real-world environments.


\subsubsection{Hyperparameters}

The hyperparameters used in the experiment are recorded in Table~\ref{tab:params}.
For VHS trigger $\eta$, the higher the value, the more the loss calculation is reflected, 
While we used the $\eta$ that we found during hyperparameter tuning, we believe that it can be set higher depending on the experimental environment.
The sampling size for PG embedding was set according to the hardware used for training, and other values were set similarly or identically to \cite{kim2023sa}.


\setlength{\tabcolsep}{4pt}
\begin{table}[h]
\begin{center}
\begin{tabular}{p{4.5cm}|p{2.8cm}}
\hline
Parameter Name    &   Value \\ \hline
Probability of VHS Trigger $\eta$ & 0.8 \\ 
Samples for PG Embedding & 64 \\
Temperature of SupCon $\tau_s$  & 0.07 \\
Warmup & 150\\
Batch Size for SupCon &   64  \\
SupCon Loss Coefficient $\eta$    &   0.5  \\
Discount $\gamma$    &   0.99  \\
Optimizer    &   Adam \\
AMSgrad    &   True  \\
Learning Rate    &   1e-4  \\
Clip Gradient Norm   &   10.0  \\
Entropy Coefficient    &   0.01  \\
Number of Training Processes    &   5  \\
Number of Test Processes & 1\\
Backpropagation Through Time & End of Episode\\
Non-Linearity   & ReLU\\
\hline
\end{tabular}
\end{center}
\caption{Hyperparameters used in our experiments.}
\label{tab:params}
\end{table}

\end{document}